\documentclass{article}
\usepackage[final]{neurips_2019}

\usepackage[utf8]{inputenc}
\usepackage[english]{babel}
\usepackage{bm,color,xcolor}
\usepackage{mathdots}
\usepackage{amsfonts}       %
\usepackage{nicefrac}       %
\usepackage{microtype}      %
\usepackage{amsmath,amsthm}
\usepackage{natbib}
\usepackage{caption}
\usepackage{booktabs} 
\usepackage{nicefrac}
\usepackage{placeins}
\usepackage{wrapfig}
\usepackage{enumitem}
\usepackage{dsfont}

\definecolor{mydarkblue}{rgb}{0,0.08,0.45}
\usepackage[colorlinks=true,
    linkcolor=mydarkblue,
    citecolor=mydarkblue,
    filecolor=mydarkblue,
    urlcolor=mydarkblue]{hyperref} 

\usepackage{todonotes}

\usepackage{amsmath,amsfonts,bm}
\usepackage{algorithm,algorithmic}

\newtheorem{theorem}{Theorem}
\newtheorem{lemma}{Lemma}
\newtheorem{proposition}{Proposition}

\newtheorem{corollary}{Corollary}

\newtheorem{assumption}{Assumption}

\newtheorem*{rep@theorem}{\rep@title}
\newcommand{\newreptheorem}[2]{%
\newenvironment{rep#1}[1]{%
 \def\rep@title{#2 \ref{##1}}%
 \begin{rep@theorem}}%
 {\end{rep@theorem}}}
\newreptheorem{theorem}{Theorem'}
\newtheorem*{rep@corollary}{\rep@title}
\newcommand{\newrepcorollary}[2]{%
\newenvironment{rep#1}[1]{%
 \def\rep@title{#2 \ref{##1}}%
 \begin{rep@corollary}}%
 {\end{rep@corollary}}}
\newreptheorem{corollary}{Corollary'}
\newtheorem*{rep@proposition}{\rep@title}
\newcommand{\newrepproposition}[2]{%
\newenvironment{rep#1}[1]{%
 \def\rep@title{#2 \ref{##1}}%
 \begin{rep@proposition}}%
 {\end{rep@proposition}}}
\newreptheorem{proposition}{Proposition'}
\def\1{\bm{1}}

\def\vepsilon{{\bm{\epsilon}}}

\def\vh{{\bm{h}}}

\def\vu{{\bm{u}}}
\def\vv{{\bm{v}}}

\def\vx{{\bm{x}}}
\def\vy{{\bm{y}}}
\def\vz{{\bm{z}}}

\def\mA{{\bm{A}}}
\def\mB{{\bm{B}}}

\def\mD{{\bm{D}}}

\def\mI{{\bm{I}}}

\def\mM{{\bm{M}}}
\def\mN{{\bm{N}}}

\def\mQ{{\bm{Q}}}

\def\mS{{\bm{S}}}

\def\mU{{\bm{U}}}
\def\mV{{\bm{V}}}
\def\mW{{\bm{W}}}
\def\mX{{\bm{X}}}
\def\mY{{\bm{Y}}}

\def\mSigma{{\bm{\Sigma}}}

\DeclareMathAlphabet{\mathsfit}{\encodingdefault}{\sfdefault}{m}{sl}
\SetMathAlphabet{\mathsfit}{bold}{\encodingdefault}{\sfdefault}{bx}{n}

\newcommand{\N}{\mathbb{N}}

\newcommand{\R}{\mathbb{R}}

\DeclareMathOperator*{\argmin}{arg\,min}

\DeclareMathOperator{\diag}{diag}
\DeclareMathOperator{\rank}{rank}
\DeclareMathOperator{\vecspan}{\text{span}}

\newlength\myindent
\title{Implicit Regularization of Discrete Gradient Dynamics in Linear Neural Networks}

\author{
Gauthier Gidel \\ Mila \& DIRO\\
Universit\'{e} de Montr\'{e}al 
\And 
Francis Bach \\
INRIA \& \' Ecole Normale Sup\' erieure\\
PSL Research University, Paris
\And 
Simon Lacoste-Julien\thanks{CIFAR fellow, Canada CIFAR AI chair \newline 
Correspondance to the first author: \texttt{<firstname>.<lastname>@umontreal.ca}
}
\\
 Mila \& DIRO \\
 Universit\'{e} de Montr\'{e}al
}

\begin{document}

\maketitle

\begin{abstract}
\noindent
When optimizing over-parameterized models, such as deep neural networks, a large set of parameters can achieve zero training error. In such cases, the choice of the optimization algorithm and its respective hyper-parameters introduces biases that will lead to convergence to specific minimizers of the objective. Consequently, this choice can be considered as an implicit regularization for the training of over-parametrized models. In this work, we push this idea further by studying the discrete gradient dynamics of the training of a two-layer linear network with the least-squares loss. Using a time rescaling, we show that, with a vanishing initialization and a small enough step size, this dynamics sequentially learns the solutions of a reduced-rank regression with a gradually increasing rank.
\end{abstract}
\section{Introduction} %
\label{sec:introduction}
When optimizing over-parameterized models, such as deep neural networks, a large set of parameters leads to a zero training error. However they lead to different values for the test error and thus have distinct generalization properties. More specifically,~\citet[Part II]{neyshabur2017implicit} argues that the choice of the optimization algorithm (and its respective hyperparameters) provides an implicit regularization with respect to its geometry: it biases the training, finding a particular minimizer of the objective.

In this work, we use the same setting as~\citet{saxe2018mathematical}: a regression problem with least-squares loss on a multi-dimensional output. Our prediction is made either by a linear model or by a two-layer linear neural network~\citep{saxe2018mathematical}. 
We extend their work which covered the \emph{continuous} gradient dynamics, to weaker assumptions as well as analyze the behavior of the \emph{discrete} gradient updates 

We show that with a vanishing initialization and a small enough step-size, the gradient dynamics of a two-layer linear neural network sequentially learns components that can be ranked according to a hierarchical structure whereas the gradient dynamics induced by the same regression problem but with a linear prediction model instead learns these components simultaneously, missing this notion of hierarchy between components. The path followed by the two-layer formulation actually corresponds to successively solving the initial regression problem with a growing low rank constraint which is also know as reduced-rank regression~\citep{izenman1975reduced}. Note that this notion of path followed by \emph{the dynamics of a whole network} is different from the notion of path introduced by~\citet{neyshabur2015path} which corresponds to a path followed \emph{inside} a fixed network, i.e., one corresponds to training dynamics whereas the other corresponds to the propagation of information inside a network.

To sum-up, in our framework, the path followed by the gradient dynamics of a two-layer linear network provides an implicit regularization that may lead to potentially better generalization properties. Our contributions are the following:
\begin{itemize}[leftmargin=1cm]
	\item Under some assumptions (see Assumption~\ref{assump:commutativity}), we prove that both the \emph{discrete} and continuous gradient dynamics sequentially learn the solutions of a gradually less regularized version of reduced-rank regression (Corollary~\ref{cor:low_rank_solution} and~\ref{cor:discrete_case}). Among the close related work, such result on implicit regularization regarding discrete dynamics is novel. For the continuous case, we weaken the standard commutativity assumption using perturbation analysis.
	\item We experimentally verify the reasonableness of our assumption and observe improvements in terms of generalization (matrix reconstruction in our case) using the gradient dynamics of the two-layer linear network when compared against the linear model.  
\end{itemize}

\subsection{Related Work} %
\label{sec:related_work}

The implicit regularization provided by the choice of the optimization algorithm has recently become an active area of research in machine learning, putting lot of interest on the behavior of gradient descent on deep over-parametrized models~\citep{neyshabur2014search,neyshabur2017geometry,zhang2017understanding}.

Several works show that gradient descent on \emph{unregularized} problems actually finds a minimum norm solution with respect to a particular norm that drastically depends on the problem of interest.
\citet{soudry2017implicit} look at a logistic regression problem and show that the predictor does converge to the max-margin solution. A similar idea has been developed in the context of matrix factorization~\citep{gunasekar2017implicit}. Under the assumption that the observation matrices commute, they prove that gradient descent on this non-convex problem finds the minimum nuclear norm solution of the reconstruction problem, they also conjecture that this result would still hold without the commutativity assumption. This conjecture has been later partially solved by~\citet{li2018algorithmic} under mild assumptions (namely  the restricted isometry property). This work has some similarities with ours, since both focus on a least-squares regression problem over matrices with a form of matrix factorization that induces a non convex landscape. Their problem is more general than ours (see~\citet{uschmajew2018critical} for an even more general setting) but they are showing a result of a different kind from ours: they focus on the properties of the limit solution the continuous dynamics whereas we show some properties on the whole dynamics (continuous \emph{and} discrete), proving that it actually visits points \emph{during} the optimization that may provide good generalization. Interestingly, both results actually share common assumptions such as a commutativity assumption (which is less restrictive in our case since it is always true in some realistic settings such as linear autoencoders), vanishing initialization and a small enough step size. 
 
\citet{nar2018step}~ also analyzed the gradient descent algorithm on a least-squares linear network model as a discrete time dynamical system, and derived certain necessary (but not sufficient) properties of the local optima that the algorithm can converge to with a non-vanishing step size. In this work, instead of looking at the properties of the limit solutions, we focus on the path followed by the gradient dynamics and precisely caracterize the weights learned along this path. 

\citet{combes2018learning} studied the continuous dynamics of some non-linear networks under relatively strong assumptions such as the linear separability of the data. Conversely, in this work, we do not make such separability assumption on the data but focus on linear networks.

Finally, \citet{gunasekar2018implicit}~compared the implicit regularization provided by gradient descent in deep linear \emph{convolutional} and \emph{fully connected} networks. They show that the solution found by gradient descent is the minimum norm for both networks but according to a different norm.
In this work, the fact that gradient descent finds the minimum norm solution is almost straightforward using standard results on least-squares. But the path followed by the gradient dynamics reveals interesting properties for generalization.
As developed earlier, instead of focusing on the properties of the solution found by gradient descent, our goal is to study the path followed by the \emph{discrete} gradient dynamics in the case of a two-layer linear network.

Prior work~\citep{saxe2013learning,saxe2014exact,advani2017high,saxe2018mathematical,lampinen2018an} studied the gradient dynamics of two-layer linear networks and proved a result similar to our Thm.~\ref{thm:eig_values}. We consider~\citet{saxe2018mathematical} as the closest related work, we re-use their notion of \emph{simple deep linear neural network}, that we call two-layer neural networks, and use some elements of their proofs to extend their results. However, note that their work comes from a different perspective: through a mathematical analysis of a simple non-linear dynamics, they intend to highlight \emph{continuous} dynamics of learning where one observes the sequential emergence of hierarchically structured notions to explain the regularities in representation of human semantic knowledge. 
In this work, we are also considering a two-layer neural network but with an optimization perspective. We are able to extend~\citet[Eq. 6 and 7]{saxe2018mathematical} weakening the commutativity assumption considered in~\citet{saxe2018mathematical} using perturbation analysis. In~\S\ref{sub:verification_of_assumption_1_for_classification_}, we test to what extent our weaker assumption holds. Our main contribution is to show a similar result on the \emph{discrete} gradient dynamics, that is important in our perspective since we aim to study the dynamics of gradient descent. This result cannot be trivially extended from the result on the \emph{continuous} dynamics. We provide details on the difficulties of the proof in \S\ref{sub:discrete_dynamics}.

\section{A Simple Deep Linear Model} %
\label{sec:a_simple_deep_linear_model}

In this work, we are interested in analyzing a least-squares model with multi-dimensional outputs. Given a \emph{finite} number $n$ of inputs $\vx_i \in \R^d \,,\; {1\leq i\leq n}$ we want to predict a \emph{multi-dimensional outputs} $\vy_i \in \R^p \,,\;{1\leq i\leq n}$
with a \emph{deep linear network}~\citep{saxe2018mathematical,gunasekar2018implicit},
\begin{equation}\label{eq:deep_model}
	\text{Deep linear model:} \quad \hat \vy^{d}(\vx) := \mW_L^\top \cdots \mW_1^\top \vx \,,
\end{equation}
where $\mW_1, \ldots, \mW_L$ are learned through a MSE formulation with the least-squares loss $f$,
\begin{equation}\label{eq:MSE_deep}
	(\mW_1^*, \ldots, \mW_L^*) \in \argmin_{\substack{\mW_l \in \mathbb{R}^{r_{l-1}\times r_{l}}\\ 1\leq l \leq L}} 	\frac{1}{2n} \|\mY -  \mX\mW_1 \cdots \mW_L  \|_2^2 =: f(\mW_1,\ldots,\mW_L)  \,,
\end{equation}
where $r_0 = d$, $r_l \in \N\,,\, 1 \leq l \leq  L-1$ and $r_L = p$, $\mX \in \mathbb{R}^{n \times d}$ and $\mY \in \mathbb{R}^{n \times p}$ are such that,
\begin{equation}
\label{eq:design_matrices}	
\mX^\top := \begin{pmatrix} 
	\vx_1 \!&\!
	\cdots \!&\!
	\vx_n
	\end{pmatrix} 
\;\; \text{and} \;\; 
\mY^\top:= \begin{pmatrix}
	\vy_1 \!&\!
	\cdots \!&\!
	\vy_n
	\end{pmatrix} \,,
\end{equation}
are the  \emph{design matrices} of $(\vx_i)_{1\leq i\leq n}$ and $(\vy_i)_{1\leq i \leq n}$. The \emph{deep linear model}~\eqref{eq:deep_model} is a $L$-layer deep linear neural network where we see $\vh_l :=  \mW_l \cdots \mW_1 \vx$ for $1\leq l \leq L-1$ as the $l^{th}$ \emph{hidden layer}. At first, since this deep linear network cannot represent more than a linear transformation, we could think that there is no reason to use a deeper representation $L=1$. However, in terms of learning flow, we will see in \S\ref{sec:gradient_flow} that for $L=2$ this model has a completely different dynamics from $L=1$.

Increasing $L$ may induce a low rank constraint when $r := \min \{r_l \,:\, 1\leq l\leq L-1\} < \min(d,p)$. In that case,~\eqref{eq:MSE_deep} is equivalent to a reduced-rank regression,
\begin{equation}\label{eq:MSE_linear_low_rank}
	\mW^{k,*} \in \argmin_{\substack{\mW \in \mathbb{R}^{p \times d}\\\rank(\mW)\leq r}} \frac{1}{2n} \sum_{i=1}^n \|\mY -  \mX\mW  \|_2^2 \,.
\end{equation}
These problems have explicit solutions depending on $\mX$ and $\mY$~\citep[Thm. 2.2]{velu2013multivariate}.

Note that, in this work we are interested in the implicit regularization provided in the context of \emph{over-parametrized models}, i.e., when $r > \min(p,d)$. In that case, 
\begin{equation}	\label{eq:quivalence_overparametrized} \notag
\{\mW_1 \cdots \mW_L \, :\, \mW_l \in \mathbb{R}^{r \times {l-1},r_l},\, 1\leq l\leq L\} = \mathbb{R}^{p \times d} \,.
\end{equation}

\section{Gradient Dynamics as a Regularizer} %
\label{sec:gradient_flow}
In this section we would like to study the \emph{discrete} dynamics of the gradient flow of~\eqref{eq:MSE_deep}, i.e., 
\begin{equation}\label{eq:deep_discrete_dynamics}
	 \mW_l^{(t+1)} = \mW_l^{(t)} - \eta \nabla_{\mW_l} f\big(\mW_{[L]}^{(t)}\big) 
	 \qquad
	 \mW_l^{(0)} \in \mathbb{R}^{r_{l-1} \times r_l} 
\, , 1\leq l\leq L \,,
\end{equation}
where we use the notation $\mW_{[L]}^{(t)} :=(\mW_1^{(t)},\ldots,\mW_L^{(t)})$. 
The quantity $\eta$ is usually called the \emph{step-size}.
In order to get intuitions on the discrete dynamics we also consider its respective \emph{continuous} version,
\begin{equation}\label{eq:deep_continuous_dynamics}
	\dot \mW_l(t) = - \nabla_{\mW_l} f\big(\mW_{[L]}(t)\big)
	\qquad
	 \mW_l(0) \in \mathbb{R}^{r_{l-1} \times r_l} 
	\, ,\; 1\leq l\leq L \,,
\end{equation}
where for $1\leq l\leq L$, $\dot \mW_l(t)$ is the temporal derivative of $\mW_l(t)$.
Note that there is no step-size in the continuous time dynamics since it actually corresponds to the limit of~\eqref{eq:deep_discrete_dynamics} when $\eta\to 0$. 
The continuous dynamics may be more convenient to study because such differential equations may have closed form solutions. In \S\ref{sub:continuous_dynamics}, we will see that under reasonable assumptions it is the case for~\eqref{eq:deep_continuous_dynamics}.

\subsection{Continuous dynamics} %
\label{sub:continuous_dynamics}

\paragraph{Linear model: $L=1$.} We start with the study of the continuous linear model, its gradient is, 
\begin{equation}\label{eq:grad_g}
	\nabla f(\mW) = \mSigma_{x}\mW - \mSigma_{xy},
\end{equation}
where $\mSigma_{xy} := \frac{1}{n} \mX^\top \mY$ and $\mSigma_{x} := \frac{1}{n} \mX^\top \mX$. Thus, $\mW(t)$ is the solution of the differential equation,
\begin{equation}\label{eq:edp_linear}
	\dot \mW(t) = \mSigma_{xy} - \mSigma_{x}\mW(t) \, , \quad \mW(0) = \mW_0 \,.
\end{equation}
\begin{proposition}\label{prop:sol_linear_edp}
For any $\mW_0 \in \mathbb{R}^{d \times p}$ , the solution to the linear differential equation~\eqref{eq:edp_linear} is 
\begin{equation}\label{eq:A_t_non_fact}
	\mW(t) = e^{-t\mSigma_x}(\mW_0-\mSigma_x^\dagger \mSigma_{xy}) + \mSigma_x^{\dagger} \mSigma_{xy} \,,
\end{equation}
where $\mSigma_x^\dagger$ is the pseudoinverse of $\mSigma_x$.
\end{proposition}
This standard result on ODE is provided in \S\ref{sub:proof_of_proposition_prop:sol_linear_edp}. 
Note that when $\mW_0 \to \bm{0}$ we have
\begin{equation}
\mW(t)\underset{\mW_0\to 0}{\to} (\mI_d- e^{-t\mSigma_x})\mSigma_x^{\dagger} \mSigma_{xy}\,.	
\end{equation}

 \paragraph{Deep linear network: $L \geq 2$.} %
 The study of the deep linear model is more challenging since for $L\geq 2$, the landscape of the objective function $f$ is non-convex. The gradient flow of~\eqref{eq:MSE_deep} is
\begin{equation}
\label{eq:grad_f}
	\nabla f_{\mW_l}(\mW_{[L]})
	=  \mW_{1:l-1}^\top (  \mSigma_x\mW - \mSigma_{xy}) \mW_{l+1:L}^\top  
	\quad \text{where} \quad \mW_{i:j}:= \mW_i \cdots \mW_j\, ,\; 1\leq l\leq L\,,
\end{equation}
where we used the convention that $\mW_{1,0} = \mI_d$ and $\mW_{L+1,L} = \mI_p$. Thus~\eqref{eq:deep_continuous_dynamics} becomes
\begin{equation}\label{eq:edp_deep}
	\dot \mW_l(t) =  \mW_{1:l-1}(t)^\top  (\mSigma_{xy}  -  \mSigma_x\mW(t)) \mW_{l+1:L}(t)^\top \,,
	\quad 
	 \mW_l(0) \in \mathbb{R}^{d \times p} \, ,\quad 1\leq l\leq L \,.
\end{equation}
We obtain a \emph{coupled} differential equation~\eqref{eq:edp_deep} that is harder to solve than the previous linear differential equation~\eqref{eq:edp_linear} due, at the same time, to its non-linear components and to the coupling between $\mW_l\,,\; 1 \leq l\leq L$. However, in the case $L=2$,~\citet{saxe2018mathematical} managed to find an explicit solution to this coupled differential equation under the assumption that ``perceptual correlation is minimal'' ($\mSigma_x = \mI_d$).\footnote{By a rescaling of the data, their proof is valid for any matrix $\mSigma_x$ \emph{proportional} to the identity matrix.} In this work we extend~\citet[Eq. 7]{saxe2018mathematical} (for $L=2$) under weaker assumptions. More precisely, we do not require the covariance matrix $\mSigma_x$ to be the identity matrix. 
Let $(\mU,\mV,\mD)$ be the SVD of $\mSigma_{xy}$, our assumption is the following:
\begin{assumption}\label{assump:commutativity}
There exist two orthogonal matrices $\mU$, $\mV$ such that we have the joint decomposition,
\begin{equation}\label{eq:decomposition_Sigmas}
	\mSigma_{x} = \mU (\mD_x + \mB) \mU^\top  \qquad \text{and} \qquad \mSigma_{xy} = \mU \mD_{xy} \mV^\top 
	\,, 
\end{equation}
where $\mB$ is such that $\|\mB\|_2 \leq \epsilon$ and $\mD_x,\,\mD_{xy}$ are matrices only with diagonal coefficients. We note $\sigma_1\geq\dots \geq\sigma_{r_{xy}}>0$ the singular values of $\mSigma_{xy}$ and $\lambda_1,\ldots,\lambda_{r_x}$ the diagonal entries of $\mD_x$. 
\end{assumption}
Since two matrices commute if and only if they are co-diagonalizable~\citep[Thm. \!1.3.21]{horn1985matrix}, the quantity $\epsilon$ represent to what extend $\mSigma_x$ and $\mSigma_{xy} \mSigma_{xy}^\top$ do not commute.
Before solving~\eqref{eq:edp_deep} under Assump.~\ref{assump:commutativity}, we describe some motivating examples where the quantity $\epsilon$ is small or zero:

\begin{itemize}[leftmargin=1cm]
	\item \textbf{Linear autoencoder}: If $\mY$ is set to $\mX$ and $L=2$, we recover a linear autoencoder: $\hat\vx (\vx) = \mW_2^\top \mW_1^\top \vx$, where $\vh := \mW_1^\top \vx$ is the \emph{encoded} representation of $\vx$, 
	\begin{equation}
		\mSigma_{xy}\mSigma_{xy}^\top = \left(\tfrac{1}{n} \mX^\top \mX \right)^2= \mSigma_x^2\,. 
		\qquad \text{Thus} ,\, \mB = 0\,.
	\end{equation}
	Note that this linear autoencoder can also be interpreted as a form of principal component analysis. Actually, if we initialize with $\mW_1 = \mW_2^\top$, the gradient dynamics exactly recovers the PCA of $\mX$, which is closely related to the matrix factorization problem of~\citet{gunasekar2017implicit}. See \S\ref{sec:deep_linear_autoencoder_recovers_pca_} where this derivation is detailed.
	\item \textbf{Deep linear $multi$-class prediction:} In that case, $p$ is the number of classes and $\vy_i$ is a one-hot encoding of the class with, in practice, $p \ll d$. 
	The intuition on why we may expect $\|\mB\|_2$ to be small is because $\rank(\mY) \ll \rank(\mX)$ and thus the matrices of interest only have to almost commute on a small space in comparison to the whole space, thus $\mB$ would be close to $0$. We verify this intuition by computing $\|\mB\|_2$ for several  classification datasets in Table~\ref{tab:assump:1}.
	\item \textbf{Minimal influence of perceptual correlation:} $\mSigma_x \approx \mI_d$. It is the setting discussed by~\citet{saxe2018mathematical}. We compare this assumption for some classification datasets with our Assump.~\ref{assump:commutativity} in~\S\ref{sub:verification_of_assumption_1_for_classification_}.
\end{itemize}
\paragraph{An explicit solution for $L=2$.} %
\label{par:explicit_solution_}
Under Assump.~\ref{assump:commutativity} and specifying the initialization, one can solve the matrix differential equation for $\epsilon =0$ and then use perturbation analysis to assess how close the solution of~\eqref{eq:edp_linear} is to the closed form solution derived for $\epsilon =0$. This result is summarized in the following theorem proved in \S\ref{sub:proof_of_theorem_1}.
\begin{theorem} \label{thm:solution continuous}
When $L=2$, under Assump.~\ref{assump:commutativity}, if we initialize with $\mW_1(0) =\mU  \diag(e^{-\delta_1},\ldots,e^{-\delta_p})  \mQ$ and $\mW_2(0) =  \mQ^{-1} \diag(e^{-\delta_1},\ldots,e^{-\delta_d}) \mV^\top $ where $\mQ$ is an arbitrary invertible matrix, then the solution of~\eqref{eq:edp_deep} can be decomposed as the sum of the solution for $\epsilon =0$ and a perturbation term,
\begin{equation}\label{eq:solution_M_N}
	\left\{
 	\begin{aligned}
 	&\mW_1(t) = \mW_1^0(t)  + \mW_1^\epsilon(t) \quad \text{where} \quad \mW_1^0(t) := \mU \diag\big(\sqrt{w_1(t)},\ldots,\sqrt{w_p(t)}\big) \mQ\\
 	&\mW_2(t) =  \mW_1^0(t) + \mW_2^\epsilon(t)  \quad \text{where} \quad \mW_2^0(t) := \mQ^{-1}\diag\big(\sqrt{w_1(t)},\ldots,\sqrt{w_d(t)}\big)\mV^\top
 	\end{aligned}
 	\right.
 \end{equation} 
 where we have $c>0$ such that $\|\mW_i^\epsilon(t)\| \leq \epsilon\cdot  e^{c t^{2}}$ and,
\begin{equation}\label{eq:m_i_ni} 
	w_i(t)  = \frac{ \sigma_i e^{2\sigma_i t - 2 \delta_i}}{ \lambda_i (e^{2\sigma_i t - 2 \delta_i} - e^{-2 \delta_i})+ \sigma_i} \,, \,1\leq i \leq r_{xy} \,,\;
	w_i(t) = \frac{e^{-2\delta_i}}{1 + 2 e^{-\delta_i} \lambda_i t} \,, \,  r_{xy} < i \leq r_x 
\end{equation}
where $(\sigma_i)$ and $(\lambda_{i})$ are defined is Assump.~\ref{assump:commutativity}. Note that $\rank(\mSigma_{xy}) := r_{xy} \leq \rank(\mSigma_x) :=r_x$.
\end{theorem}
The main difficulty in this result is the perturbation analysis for which we use a consequence of Grönwall's inequality~\citep{gronwall1919note} (Lemma~\ref{lemma:gronwall}). The proof can be sketched in three parts: first showing the result for $\epsilon =0$, then showing that in the case $\epsilon > 0$, the matrices $\mW_1(t)/t$ and $\mW_2(t)/t$ are bounded and finally use Lemma~\ref{lemma:gronwall} to get the perturbation bound.

This result is more general than the one provided by~\citet{saxe2018mathematical} because it requires a weaker assumption than $\mSigma_x = \mI_d$ and $\epsilon=0$. In doing so, we obtain a result that takes into account the influence of correlations of the input samples. Note that Thm.~\ref{thm:solution continuous} is only valid if the initialization $\mW_1(0)\mW_2(0)$ has the same singular vectors as $\mSigma_{xy}$. However, making such assumptions on the initialization is standard in the literature and, in practice, we can set the initialization of the optimization algorithm in order to also ensure that property. For instance, in the case of the linear autoencoder, one can set $\mW_1(0) = \mW_2(0) = e^{-\delta}\mI_d$.

In the following subsection we will use Thm.~\ref{thm:solution continuous} to show that the components $[\mU]_i \,,\; 1\leq i \leq r_{xy}$ in the order defined by the decreasing singular values of $\mSigma_{xy}$ are learned sequentially by the gradient dynamics.

\paragraph{Sequential learning of components.} %
\label{sub:implicit_regularization}

The sequential learning of the left singular vectors of $\mSigma_{xy}$ (sorted by the magnitude of its singular values) by the \emph{continuous} gradient dynamics of deep linear networks has been highlighted by~\citet{saxe2018mathematical}. They note in their Eq.~(10) that the $i^{th}$ phase transition happens approximately after a time $T_i$ defined as (using our notation),
\begin{equation}\label{eq:phase transition}
	T_i := \frac{\delta_i}{\sigma_i} \ln(\sigma_i) \quad \text{where}\quad \mSigma_{xy} = \sum_{i=1}^{r_{xy}} \sigma_i \vu_i \vv_i^\top \,.
\end{equation}
They argue that as $\delta_i \to \infty$, the time $T_i$ is roughly $O(1/\sigma_i)$. The intuition is that a vanishing initialization increases the gap between the phase transition times $T_i$ and thus tends to separate the learning of each components. However, a vanishing initialization just formally leads to $T_i \to \infty$.

In this work, we introduce a notion of \emph{time rescaling} in order to formalize this notion of phase transition and we show that, after this time rescaling, the point visited between two phase transitions is the solution of a low rank regularized version~\eqref{eq:MSE_linear_low_rank} of the initial problem~\eqref{eq:MSE_deep} with the low rank constraint that loosens sequentially. 

The intuition behind time rescaling is that it counterbalances the vanishing initialization in~\eqref{eq:phase transition}: Since $T_i$ grows as fast as $\delta_i$ we need to multiply the time by $\delta_i$, in order to grow at the same pace as $T_i$.

\begin{wrapfigure}{r}{7cm}
\vspace{-4mm}
\centering
\includegraphics[width = .95\linewidth]{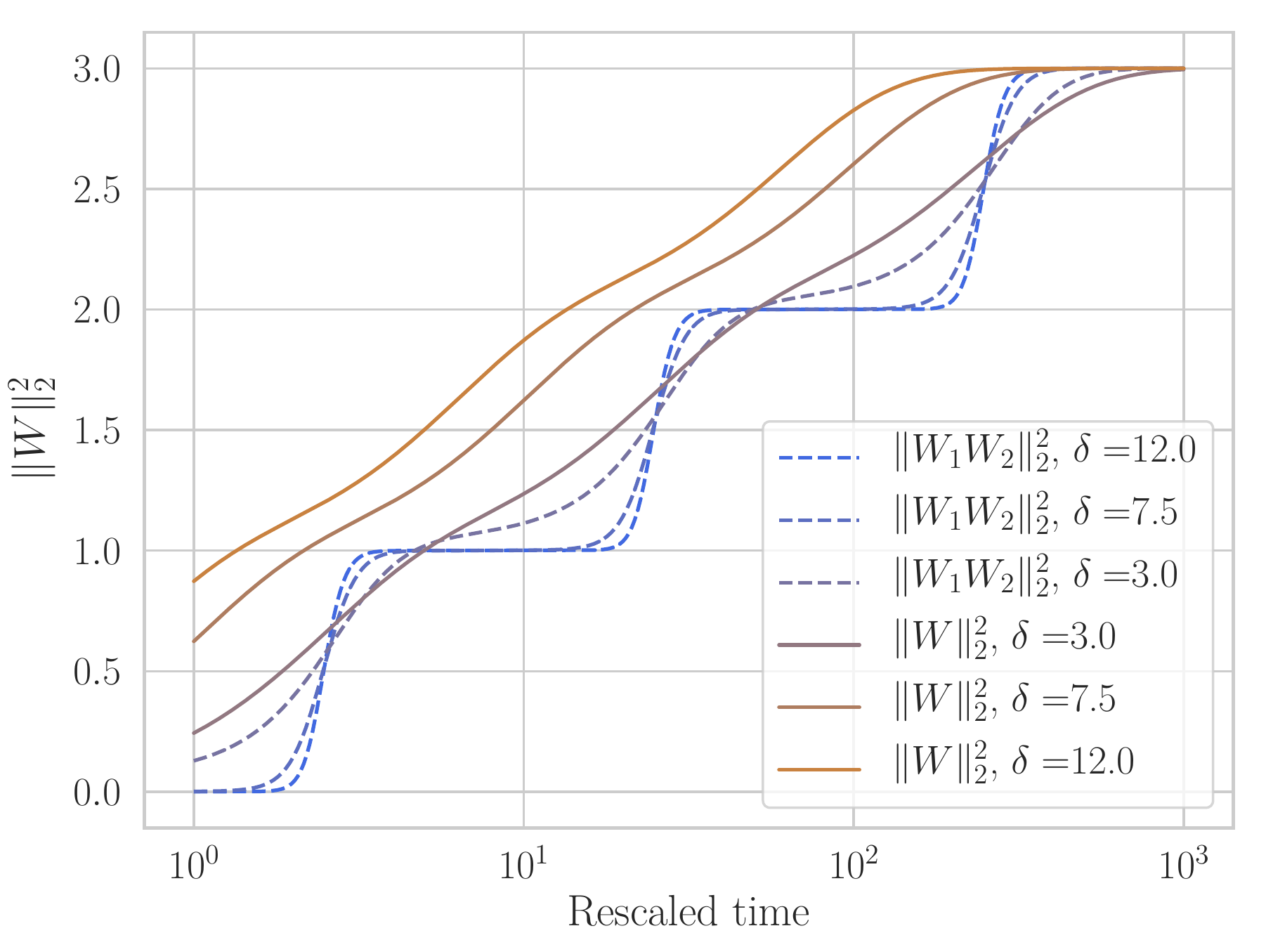}
\caption{\small
Closed form solution of squared $\ell_2$ norm of $\mW(\delta t)$ and $\mW_1(\delta t) \mW_2(\delta t)$ respectively for a linear model and a two-layer linear autoencoder, depending on $\mW(0) = \mW_1(0) \mW_2(0)= e^{-\delta}\mI_d$. Note that for an autoencoder $\lambda_i = \sigma_i$ and thus the trace norm has integer values.
According to Thm.~\ref{thm:eig_values}, the integer trace norm increment represents the learning of a new component.}
\vspace{-2mm}
\label{fig:trace_norm_init}
\end{wrapfigure}

Using this rescaling we can present our theorem, proved in \S\ref{sub:proof_of_theorem_eig_vlaues}, which says that a vanishing initialization tends to force the sequential learning of the component of $\mX$ associated with the largest singular value of $\mSigma_{xy}$. Note that we need to rescale the time \emph{uniformly} for each component. That is why in the following we set $\delta_i= \delta \,,\,1\leq i\leq \max(p,d)$.
\begin{theorem}\label{thm:eig_values} Let us denote $w_i(t)$, the values defined in~\eqref{eq:m_i_ni}. If $w_i(0) = e^{-\delta}\,,\;1\leq i\leq r,$ and $\epsilon = e^{-\delta^{2}\ln(\delta)}$ then we have that $w_i(\delta t)$ converge to a step function as $\delta \to \infty$:
\begin{equation}
	 w_i(\delta t) \underset{\delta \to \infty}{\to} 
	\tfrac{\sigma_i}{\lambda_i + \sigma_i}\mathds{1}\{t = T_i \} + \tfrac{\sigma_i}{\lambda_i}\mathds{1}\{t > T_i \} \,.
\end{equation}
where $T_i:=1/\sigma_i$, $\mathds{1}\{t\in A\} = 1$ if $\,t \in A$ and $0$ otherwise.  
\end{theorem}
Notice how the $i^{th}$ components of $\mW_1$ and $\mW_2$ are inactive, i.e., $w_i(t)$ is zero, for small $t$ and is suddenly learned when $t$ reaches the phase transition time $T_i := 1/\sigma_i$.
As shown in Prop.~\ref{prop:sol_linear_edp} and illustrated in Fig.~\ref{fig:trace_norm_init}, this sequential learning behavior does not occur for the non-factorized formulation.~\citet{gunasekar2017implicit} observed similar differences between their factorized and not factorized formulations of matrix regression.
Note that, the time rescaling we introduced is $t \rightarrow \delta t$, in order to compensate the vanishing initialization, rescaling the time and taking the limit this way for~\eqref{eq:edp_linear} would lead to a constant function. 

\citet{gunasekar2017implicit} also had to consider a vanishing initialization in order to show that on a simple matrix factorization problem the continuous dynamics of gradient descent does converge to the minimum nuclear norm solution. This assumption is necessary in such proofs in order to avoid to initialize with wrong components. However one cannot consider an initialization with the null matrix since it is a stationary point of the dynamics, that is why this notion of double limit (vanishing initialization and $t\to \infty$) is used.

From Thm.~\ref{thm:eig_values}, two corollaries follow directly. The first one regards the nuclear norm of the product $\mW_1(\delta t)\mW_2(\delta t)$. This corollary says that $\|\mW_1(\delta t)\mW_2(\delta t)\|_*$ is a step function and that each increment of this integer value corresponds to the learning of a new component of $\mX$. These components are leaned by order of relevance, i.e., by order of magnitude of their respective eigenvalues and the learning of a new component can be easily noticed by an incremental gap in the nuclear norm of the matrix product $\mW_1(\delta t)\mW_2(\delta t)$,
\begin{corollary}\label{cor:trace_rank}
Let $\mW_1(t)$ and $\mW_2(t)$ be the matrices solution of~\eqref{eq:edp_deep} defined in~\eqref{eq:solution_M_N}. The limit of the squared euclidean norm of $\mW_1(t)\mW_2(t)$ when $\delta \to \infty$ is a step function defined as, 
\begin{equation}
	\|\mW_1(\delta t)\mW_2(\delta t)\|_2^2 \underset{\delta \to \infty}{\to} 
	\sum_{i=1}^{r_{xy}} \tfrac{\sigma_i^2}{\lambda_i^2}
	\mathds{1}\{T_i < t\}
	+\tfrac{\sigma_i^2}{(\lambda_i+\sigma_i)^2}
	\mathds{1}\{T_i = t \}
\end{equation}
where $T_i:=1/\sigma_i$ and $\sigma_1 > \cdots > \sigma_{r_{xy}} > 0$ are the positive singular values of $\mSigma_{xy}$. 
\end{corollary}

It is natural to look at the norm of the product $\mW_1(\delta t)\mW_2(\delta t)$ since in Thm.~\ref{thm:eig_values}, $(w_i(t))$ are its singular values. However, since the rank of $\mW_1(\delta t)\mW_2(\delta t)$ is discontinuously increasing after each phase transition, any norm would jump with respect to the rank increments.
We illustrate these jumps in Fig.~\ref{fig:trace_norm_init} where we plot the closed form of the squared $\ell_2$ norms of $t \mapsto \mW(\delta t)$ and $ t \mapsto \mW_1(\delta t)\mW_2(\delta t)$ for vanishing initializations with $\mSigma_{yx} = \diag(10^{-1},10^{-2},10^{-3})$ and $\mSigma_{x} = \mI_d$.

From Thm.~\ref{thm:eig_values}, we can notice that, between time $T_k$ and $T_{k+1}$, the rank of the limit matrix $\mW_1\mW_2$ is actually equal to $k$, meaning that at each phase transition, the rank of $\mW_1\mW_2$ is increased by 1. Moreover, this matrix product contains the $k$ components of $\mX$ corresponding to the $k$ largest singular values of $\mSigma_{xy}$. Thus, we can show that this matrix product is the solution of the $k$-low rank constrained version~\eqref{eq:MSE_linear_low_rank} of the initial problem~\eqref{eq:MSE_deep},
\begin{corollary}\label{cor:low_rank_solution}
Let $\mW_1(t)$ and $\mW_2(t)$ be the matrices solution of~\eqref{eq:edp_deep} defined in~\eqref{eq:solution_M_N}. We have that,
\begin{equation}
	\tfrac{1}{\sigma_k}<t<\tfrac{1}{\sigma_{k+1}} \quad  \Rightarrow  \quad
	\mW_1(\delta t)\mW_2(\delta t)  \underset{\delta \to \infty}{\to} \mW^{k,*}   \,,   \qquad 1\leq k \leq r_{xy}\,.
\end{equation}
where $\mW^{k,*}$ is the minimum $\ell_2$ norm solution of the reduced-rank-$k$ regression problem~\eqref{eq:MSE_linear_low_rank} .
\end{corollary}

\subsection{Discrete dynamics} %
\label{sub:discrete_dynamics}
We are interested in the behavior of optimization methods. Thus, the gradient dynamics of interest is the \emph{discrete} one~\eqref{eq:deep_discrete_dynamics}. A major contribution of our work is thus contained in this section.
The continuous case studied in \S~\ref{sub:continuous_dynamics} provided good intuitions and insights on the behavior of the potential discrete dynamics that we can use for our analysis.

\paragraph{Why the discrete analysis is challenging.}
Previous related work~\citep{gunasekar2017implicit,saxe2018mathematical} only provide results on the continuous dynamics. Their proofs use the fact that their respective continuous dynamics of interest have a closed form solution (e.g., Thm.\ref{thm:solution continuous}). To our knowledge, no closed form solution is known for the discrete dynamics~\eqref{eq:deep_discrete_dynamics}. Thus its analysis requires a new proof technique.
Moreover, using Euler’s integration methods, one can show that both dynamics are close but only for a vanishing step size depending on a \emph{finite} horizon. Such dependence on the horizon is problematic since the time rescaling used in Thm.~\ref{thm:eig_values} would make any finite horizon go to infinity.
In this section, we consider realistic conditions on the step-size (namely, it has to be smaller than the Lipschitz constant and some notion of eigen-gap) without any dependence on the horizon.
Such assumption is relevant since we want to study the dynamics of practical optimization algorithms (i.e., with a step size as large as possible and without knowing in advance the horizon).

\paragraph{Single layer linear model.} %
\label{par:non_factorized_pca}
In this paragraph, we consider the discrete update for the linear model. Since $L=1$, for notational compactness, we call $\mW_t$ the matrix updated according to~\eqref{eq:deep_discrete_dynamics}.
Using the gradient derivation~\eqref{eq:grad_g}, the discrete update scheme for the \emph{linear model} is,
\begin{equation}\label{eq:explicit_discrete_linear} \notag
	\mW_{t+1} = \mW_t - \eta(\mSigma_{x}\mW_t- \mSigma_{xy}) = (\mI_d-\eta \mSigma_{x})\mW_t + \eta \mSigma_{xy} \,.
\end{equation}
Noticing that for $1/\lambda_{\max}(\mSigma_{x})>\eta>0\,,\; \mI_d-\eta \mSigma_{x}$ is invertible, this recursion (see \S\ref{sub:proof_of_eq_eq:result_explicit}) leads to, 
\begin{equation}\label{eq:result_explicit}
	\mW_t= (\mW_0-\mSigma_x^\dagger \mSigma_{xy})(\mI_d-\eta \mSigma_{x})^{t} +\mSigma_x^\dagger\mSigma_{xy}\,.
\end{equation}
We obtain a similar result as the solution of the differential equation given in Prop.~\ref{prop:sol_linear_edp}. With a vanishing initialization we reach a function that does not sequentially learn some components.

\paragraph{Two-layer linear model.} %
\label{par:factorized_pca_}
The discrete update scheme for the \emph{two-layer linear network}~\eqref{eq:MSE_deep} is,
\begin{equation} \label{eq:explicit_discrete_deep_dynamics}
	\mW_1^{(t+1)} = \mW_1^{(t)}\! - \eta  (\mSigma_{x}\mW^{(t)}\! - \mSigma_{xy}) (\mW_2^{(t)})^\top \,, \;
	\mW_2^{(t+1)} = \mW_2^{(t)}\! - \eta  (\mW_1^{(t)})^\top (  \mSigma_x\mW^{(t)} \! - \mSigma_{xy}) \, .  
\notag
\end{equation}
When $\epsilon=0$, by a change of basis and a proper initialization, we can reduce the study of this matrix equation to $r$ independant dynamics (see \S\ref{sub:proof_of_theorem_3} for more details), for $1\leq i \leq r$, 
\begin{equation} \label{eq:rec_coeffs}
	w_i^{(t+1)} = w_i^{(t)} + \eta w_i^{(t)}(\sigma_i - \lambda_i w_i^{(t)} w_i^{(t)}) \,.
\end{equation}
Thus we can derive a bound on the iterate $w_i^{(t)}$ 
leading to the following theorem,

\begin{theorem}\label{thm:discrete_case} Under the same assumptions as Thm.~\ref{thm:solution continuous} and $\epsilon=0$, we have
\begin{equation}\label{eq:solution_M_N_discrete}
 	\mW_1^{(t)} = \mU \diag\Big(\sqrt{w_1^{(t)}},\ldots,\sqrt{w_p^{(t)}}\Big) \mQ 
 	\quad \text{and} \quad 
 	\mW_2^{(t)} = \mQ^{-1}\diag\Big(\sqrt{w_1^{(t)}},\ldots,\sqrt{w_d^{(t)}}\Big)\mV^\top \,.
 	\notag
 \end{equation} 
Moreover, for any $1\leq i \leq r_{xy}$, if $1>w_i^{(0)} >0$ and $2\eta \sigma_i<1$, then $ \forall t \geq 0 \,,\; 1 \leq i \leq r_x $ we have,
\begin{equation}
	 \frac{w_i^{(0)}}{(\sigma_i- \lambda_i w_i^{(0)})e^{(-2 \eta \sigma_i  {\color{red}+ 4} \eta^2 \sigma_i^2 )t } + w_i^{(0)} \lambda_i}
	 \leq
	w_i^{(t)} 
	\leq  
	\frac{w_i^{(0)}}{(\sigma_i- \lambda_i w_i^{(0)})e^{(-2 \eta \sigma_i  {\color{red}-}  \eta^2 \sigma_i^2 )t } + w_i^{(0)} \lambda_i} \,,
	\label{eq:mu_asym_lowerbound}
\end{equation}
and $w_i^{(t)} \leq  \tfrac{w_i^{(0)}}{1 + w_i^{(0)} \lambda_i \eta t}$ for $r_{xy} \leq i \leq r_x$. The differences with the continuous case~\eqref{eq:m_i_ni} are in red.
\end{theorem}
\proof[Proof sketch] The solution of the continuous dynamics lets us think directly studying the sequence $w_i^{(t)}$ might be quite challenging since the solution of the continuous dynamics $w_i(t)^{-1}$ has a non-linear and non-convex behavior.
 
 The main insight from this proof is that one can treat the discrete case using the right transformation, to show that some sequence doee converge linearly.

Thm.~\ref{thm:eig_values} indicates the quantity $w_i(t)^{-1} - \tfrac{\lambda_i}{\sigma_i}$ is the good candidate to show linear convergence to 0,
\begin{equation}
	w_i(t)^{-1} - \tfrac{\sigma_i}{\lambda_i} = (w_i(0)^{-1} - \tfrac{\sigma_i}{\lambda_i})e^{-2 \eta \sigma_i t} \,.
\end{equation}
What we can expect is thus to show that the sequence $(w_i^{(t)})^{-1} - \tfrac{\sigma_i}{\lambda_i}$ has similar properties. 
The first step of the proof is to show that $(w_i^{(t)})$ is an increasing sequence smaller than one. The second step is then to use~\eqref{eq:rec_coeffs} to get,
\begin{align} \notag
	\tfrac{1}{w_i^{(t+1)}}-\tfrac{\lambda_i}{\sigma_i}
	& = \tfrac{1}{w_i^{(t)}} \left( \tfrac{1}{1 + 2 (\sigma_i - \lambda_i w_i^{(t)}) +  \eta^2 (\sigma_i- \lambda_i w_i^{(t)})^2} \right)   -\tfrac{\lambda_i}{\sigma_i}
\end{align}
Then using that $1-x \leq \frac{1}{1+x} \leq 1-x +x^2$ for any $1\leq x\leq 0$ we can derive the upper and lower bounds on the linear convergence rate.  See \S\ref{sub:proof_of_theorem_3} for full proof.
\endproof 
In order to get a similar interpretation of Thm.~\ref{thm:discrete_case} in terms of implicit regularization, we use the intuitions from Thm.~\ref{thm:eig_values}. The analogy between continuous and discrete time is that the discrete time dynamics is doing $t$ time-steps of size $\eta$, meaning that we have $\mW(\eta t) \approx \mW_t$, the time rescaling in continuous time consists in multiplying the time by $\delta$ thus we get the analog phase transition time,
\begin{equation}
	\eta T_i := \tfrac{1}{\sigma_i} \quad \Rightarrow \quad T_i := \tfrac{1}{\eta \sigma_i}  \,.
\end{equation}Recall that we assumed that $m_i^{(0)} = n_i^{(0)} = e^{-\delta}$. Thus, we show that the $i^{th}$ component is learned around time $T_{i}$, and consequently that the components are learned sequentially, 
\begin{corollary}\label{cor:discrete_case} 
If $\eta < \frac{1}{2\sigma_1}$, $\eta < 2\tfrac{\sigma_{i}- \sigma_{i+1}}{\sigma_i^2}$ and $\eta < \tfrac{\sigma_{i}- \sigma_{i+1}}{2\sigma_{i+1}^2}, $ for $ \;1\leq i \leq r_{xy} -1$, then for $1 \leq i < r_{x}$, 
\begin{equation}
	w_{i}^{(\delta T_j)}  \underset{\delta \to \infty}{\to} \left\{ \begin{aligned}& 0  \quad \text{if} \quad  i > r_{xy}\quad \text{or} \quad  j <i  \\
	&\frac{\sigma_i}{\lambda_i}  \quad \text{if} \quad  i \leq r_{xy} \; \,\; \text{and} \;\;\,  j > i \,.
	\end{aligned}
	\right. \:	
\end{equation}
where $T_0 := 0, \, T_j := \frac{1}{\sigma_j \eta} \,, \;  1 \leq j \leq r_{xy}$ and $T_j := +\infty \:  \text{if} \: j > r_{xy}$.
\end{corollary}
This result is proved in~\S\ref{sub:proof_of_theorem_3}. The quantities $\frac{\sigma_{i}- \sigma_{i+1}}{\sigma_i^2}$ and $\frac{\sigma_{i}- \sigma_{i+1}}{\sigma_{i+1}^2}$ can be interpreted as relative \emph{eigen-gaps}. Note that they are well defined since we assumed that the eigenspaces were unidimensional. The intuition behind this condition is that the step-size cannot be larger than the eigen-gaps because otherwise the discrete optimization algorithm would not be able to distinguish some components.

\section{Experiments} %
\label{sec:experiments}
\begin{figure}
\vspace{-8mm}
\begin{minipage}{.64\linewidth}
\hspace{-3mm}
\begin{figure}[H]
\hspace{-3mm}
\centering
\includegraphics[width = .50\linewidth]{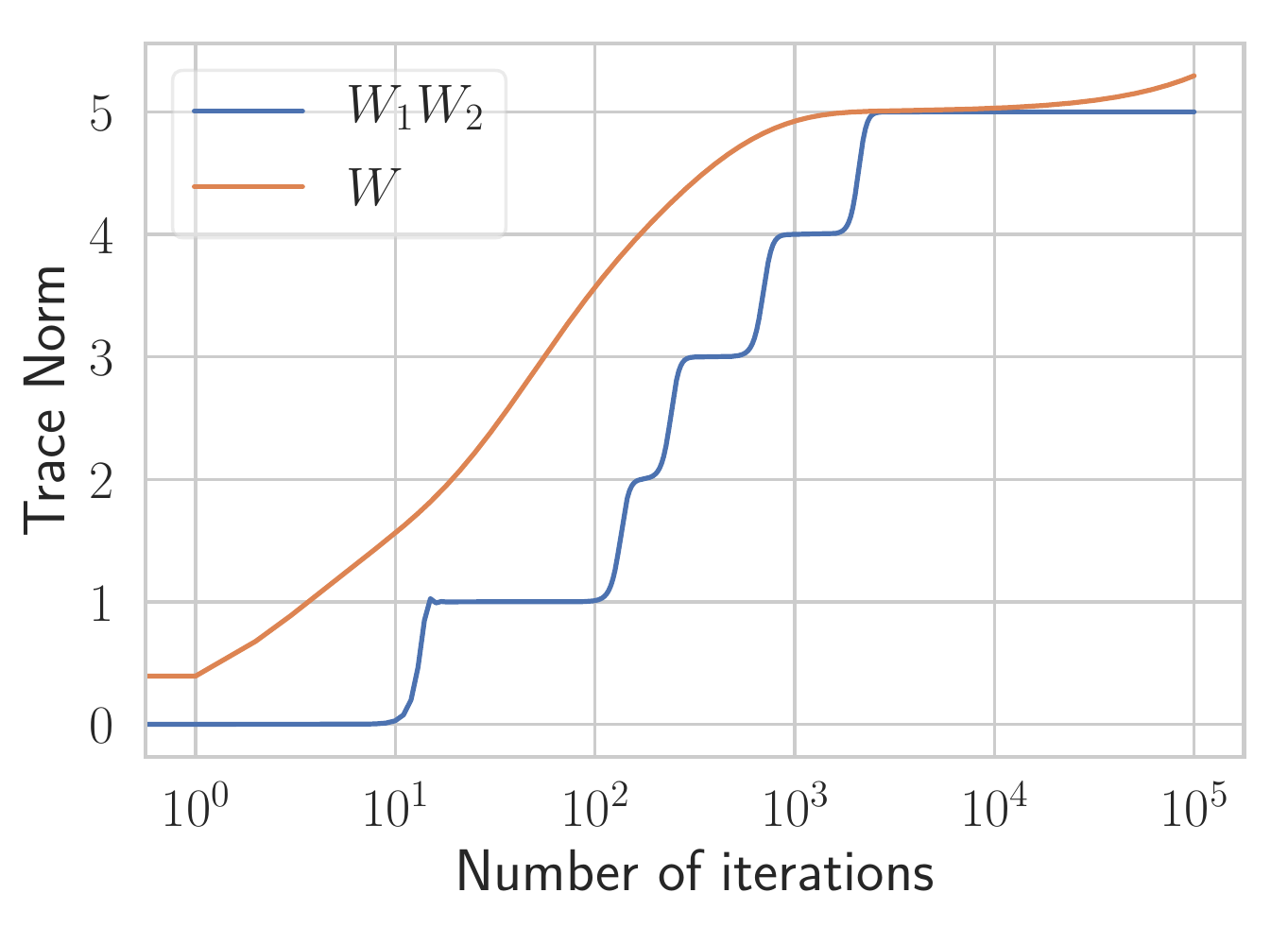}
\hspace{-3mm}
\includegraphics[width = .52\linewidth]{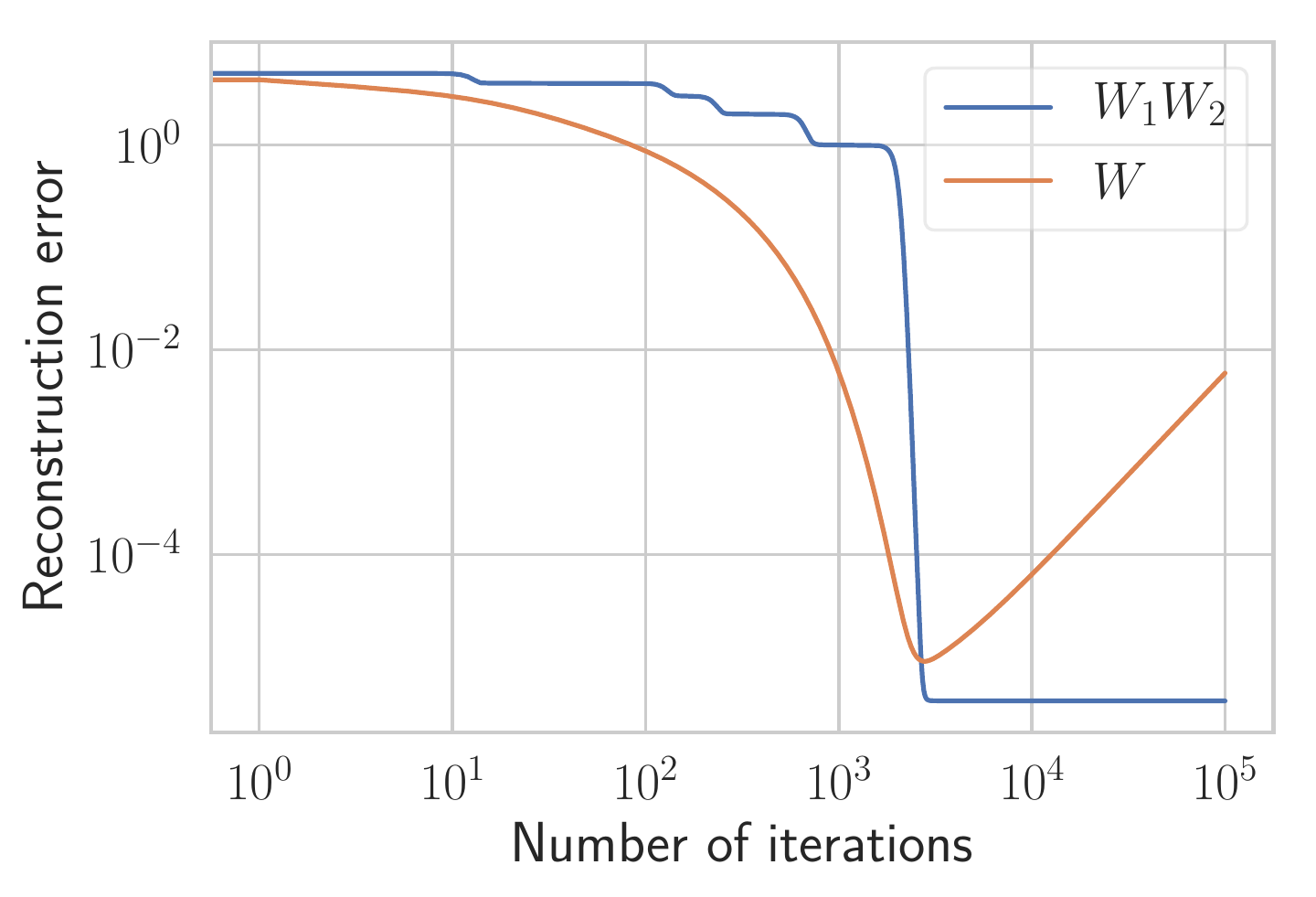}
\hspace{-3mm}
\vspace{-2mm}
\caption{ \small
Trace norm and reconstruction errors of $\mW^{(t)}$ for $L=1$ and $2$ as a function of $t$.}
\label{fig:trace_norm_exp}
\end{figure}
\end{minipage}
\hspace{.5mm}
\scalebox{.97}{
\begin{minipage}{.31\linewidth}
\begin{table}[H]
    \begin{tabular}{ccc}
      \toprule 
      Dataset &$\Delta_{xy}$ & $\Delta_{x}$  \\
      \midrule
      MNIST  & $2.8 \times 10^{-2}$ & $.70$ \\
       CIFAR-10 & $3.0 \times 10^{-2}$&$.68$\\
       ImageNet& $1.7\times 10^{-1}$ & $.70$\\
      \bottomrule
    \end{tabular}
    \vspace{8mm}
    \caption{ \small Value of the quantities $\Delta_{xy}$ and $\Delta_x$ defined in~\eqref{eq:ratio}.}
    \label{tab:assump:1}
 \vspace*{-8mm}
\end{table}
\end{minipage}
}
\end{figure}
\subsection{Assump. 1 for Classification Datasets} %
\label{sub:verification_of_assumption_1_for_classification_}
In this section we verify to what extent Assump.~\ref{assump:commutativity} is true on standard classification datasets. For this, we compute the normalized quantities $\Delta_{xy}$ and $\Delta_{x}$ representing how much Assump.~\ref{assump:commutativity} and the assumption that $\mSigma_x \approx \mI_d$ are respectively broken. We compute $\mB$ by computing $\mU$, the left singular vector of $\mSigma_{xy}$ and looking at the non-diagonal coefficients of $\mU^\top \mSigma_{x}\mU$,
\begin{align}\label{eq:ratio}
	 &\Delta_{xy}:= \tfrac{\|\mB\|_2}{\|\mSigma_x\|_2} \,,\quad
	 \Delta_{x} := \tfrac{1}{2} \big\|\hat \mSigma_x - \hat\mI_d\big\|_2 \,,
\end{align}
where $\|\cdot\|$ is the Frobenius norm, the $\hat \mSigma$ expressions correspond to $\hat \mX := \mX / \|\mX\| $ and $\hat \mI_d := \mI_d /\|\mI_d\|$. These normalized quantities are between $0$ and $1$. The closer to $1$, the less the assumption hold and conversely, the closer to $0$, the more the assumption approximately holds. We present the results on three standard classification datasets, MNIST~\citep{lecun2010mnist}, CIFAR10~\citep{krizhevsky2014cifar} and ImageNet~\citep{deng2009imagenet}, a down-sampled version of ImageNet with images of size $64 \times 64$. In Table~\ref{tab:assump:1}, we can see that the quantities $\Delta_x$ and $\Delta_{xy}$ do not vary much among the datasets and that the $\Delta$ associated with our our Assump.~\ref{assump:commutativity} is two orders of magnitude smaller than the $\Delta$ associated with~\citet{saxe2018mathematical}'s assumption indicating the relevance of our assumption.

\subsection{Linear Autoencoder} %
\label{sub:deep_linear_autoencoder_}
For an auto-encoder, we have, $\mY = \mX$.
We want to compare the reconstruction properties of $\mW^{(t)}$ computed though~\eqref{eq:explicit_discrete_linear} and of the matrix product $\mW_1^{(t)} \mW_2^{(t)}$ where $\mW_1^{(t)}$ and $\mW_2^{(t)}$ are computed though~\eqref{eq:explicit_discrete_deep_dynamics}.
In this experiment, we have $p=d=20, n=1000, r =5$ and we generated synthetic data. First we generate a \emph{fixed} matrix $\mB \in\mathbb{R}^{d\times r}$ such that, $\mB_{kl} \sim \mathcal{U}([0,1]),\, 1\leq k,l \leq n$. Then, for $1\leq i\leq n$,  we sample $\vx_i \sim \mB \vz_i + \vepsilon_i$ where $\vz_i \sim \mathcal{N}(\bm{0},\mD :=\diag(4,2,1,1/2,1/4))$ and $\vepsilon_i \sim  10^{-3}\mathcal{N}(\bm{0},\mI_d)$. In Fig.~\ref{fig:trace_norm_exp}, we plot the trace norm of $\mW^{(t)}$ and $\mW_1^{(t)} \mW_2^{(t)}$ as well as their respective reconstruction errors as a function of $t$ the number of iterations, 
\begin{equation}
\|\mW^{(t)}-\mB \mD \mB^\top \|_2 	\,.
\end{equation}
We can see that the experimental results are very close to the theoretical behavior predicted with the continuous dynamics in Figure~\ref{fig:trace_norm_init}. Contrary to the dynamics induced by the linear model formulation ($L=1$), the dynamics induced by the two-layer linear network ($L=2$) is very close to a step function: each step corresponds to the learning to a new component: They are learned sequentially. 

\section{Discussion} %
\label{sec:discussion}
There is a growing body of empirical and theoretical evidence that the implicit regularization induced by gradient methods is key in the training of deep neural networks. Yet, as noted by~\citet{zhang2017understanding}, even for linear models, our understanding of the origin of generalization is limited. 
In this work, we focus on a simple non-convex objective that is parametrized by a two-layer linear network. In the case of linear regression we show that the \emph{discrete} gradient dynamics also visits points that are implicitly regularized solutions of the initial optimization problem. 
In that sense, in the context of machine learning, applying gradient descent on the overparametrized model of interest, provides a form of implicit regularization: it sequentially learns the hierarchical components of our problem which could help for generalization.
Our setting does not pretend to solve generalization in deep neural networks; many majors components of the standard neural network training are omitted such as the non-linearities, large values of $L$ and the stochasticity in the learning procedure (SGD). Nevertheless, it provides useful insights about the source of generalization in deep learning.

\subsection*{Acknowledgments.} %
\label{par:paragraph_name}
This research was partially supported by the Canada CIFAR AI Chair Program, the Canada Excellence Research Chair in “Data Science for Realtime Decision-making”, by the NSERC Discovery Grant RGPIN-2017-06936, by a graduate Borealis AI fellowship and by a Google Focused Research award. 
\bibliographystyle{abbrvnat}
\bibliography{implicit_reg}

\begin{thebibliography}{26}
\providecommand{\natexlab}[1]{#1}
\providecommand{\url}[1]{\texttt{#1}}
\expandafter\ifx\csname urlstyle\endcsname\relax
  \providecommand{\doi}[1]{doi: #1}\else
  \providecommand{\doi}{doi: \begingroup \urlstyle{rm}\Url}\fi

\bibitem[Advani and Saxe(2017)]{advani2017high}
M.~S. Advani and A.~M. Saxe.
\newblock High-dimensional dynamics of generalization error in neural networks.
\newblock \emph{arXiv preprint arXiv:1710.03667}, 2017.

\bibitem[Berglund(2001)]{berglund2001perturbation}
N.~Berglund.
\newblock Perturbation theory of dynamical systems.
\newblock \emph{arXiv preprint math/0111178}, 2001.

\bibitem[Coddington and Levinson(1955)]{coddington1955theory}
E.~A. Coddington and N.~Levinson.
\newblock \emph{Theory of Ordinary Differential Equations}.
\newblock Tata McGraw-Hill Education, 1955.

\bibitem[Combes et~al.(2018)Combes, Pezeshki, Shabanian, Courville, and
  Bengio]{combes2018learning}
R.~T.~d. Combes, M.~Pezeshki, S.~Shabanian, A.~Courville, and Y.~Bengio.
\newblock On the learning dynamics of deep neural networks.
\newblock \emph{arXiv preprint arXiv:1809.06848}, 2018.

\bibitem[Deng et~al.(2009)Deng, Dong, Socher, Li, Li, and
  Fei-Fei]{deng2009imagenet}
J.~Deng, W.~Dong, R.~Socher, L.-J. Li, K.~Li, and L.~Fei-Fei.
\newblock Imagenet: A large-scale hierarchical image database.
\newblock In \emph{CVPR}, 2009.

\bibitem[Gronwall(1919)]{gronwall1919note}
T.~H. Gronwall.
\newblock Note on the derivatives with respect to a parameter of the solutions
  of a system of differential equations.
\newblock \emph{Annals of Mathematics}, 1919.

\bibitem[Gunasekar et~al.(2017)Gunasekar, Woodworth, Bhojanapalli, Neyshabur,
  and Srebro]{gunasekar2017implicit}
S.~Gunasekar, B.~E. Woodworth, S.~Bhojanapalli, B.~Neyshabur, and N.~Srebro.
\newblock Implicit regularization in matrix factorization.
\newblock In \emph{NIPS}, 2017.

\bibitem[Gunasekar et~al.(2018)Gunasekar, Lee, Soudry, and
  Srebro]{gunasekar2018implicit}
S.~Gunasekar, J.~Lee, D.~Soudry, and N.~Srebro.
\newblock Implicit bias of gradient descent on linear convolutional networks.
\newblock \emph{arXiv preprint arXiv:1806.00468}, 2018.

\bibitem[Horn et~al.(1985)Horn, Horn, and Johnson]{horn1985matrix}
R.~A. Horn, R.~A. Horn, and C.~R. Johnson.
\newblock \emph{Matrix Analysis}.
\newblock Cambridge University Press, 1985.

\bibitem[Izenman(1975)]{izenman1975reduced}
A.~J. Izenman.
\newblock Reduced-rank regression for the multivariate linear model.
\newblock \emph{Journal of Multivariate Analysis}, 1975.

\bibitem[Krizhevsky et~al.(2014)Krizhevsky, Nair, and
  Hinton]{krizhevsky2014cifar}
A.~Krizhevsky, V.~Nair, and G.~Hinton.
\newblock The {CIFAR-10} dataset.
\newblock \emph{online: http://www. cs. toronto. edu/kriz/cifar. html}, 2014.

\bibitem[Lampinen and Ganguli(2019)]{lampinen2018an}
A.~K. Lampinen and S.~Ganguli.
\newblock An analytic theory of generalization dynamics and transfer learning
  in deep linear networks.
\newblock In \emph{ICLR}, 2019.

\bibitem[LeCun et~al.(2010)LeCun, Cortes, and Burges]{lecun2010mnist}
Y.~LeCun, C.~Cortes, and C.~Burges.
\newblock {MNIST} handwritten digit database.
\newblock \emph{AT\&T Labs [Online]. Available: http://yann. lecun.
  com/exdb/mnist}, 2010.

\bibitem[Li et~al.(2018)Li, Ma, and Zhang]{li2018algorithmic}
Y.~Li, T.~Ma, and H.~Zhang.
\newblock Algorithmic regularization in over-parameterized matrix sensing and
  neural networks with quadratic activations.
\newblock In \emph{Conference On Learning Theory}, pages 2--47, 2018.

\bibitem[Nar and Sastry(2018)]{nar2018step}
K.~Nar and S.~Sastry.
\newblock Step size matters in deep learning.
\newblock In \emph{NeurIPS}, 2018.

\bibitem[Neyshabur(2017)]{neyshabur2017implicit}
B.~Neyshabur.
\newblock \emph{Implicit Regularization in Deep Learning}.
\newblock PhD thesis, TTIC, 2017.

\bibitem[Neyshabur et~al.(2015{\natexlab{a}})Neyshabur, Salakhutdinov, and
  Srebro]{neyshabur2015path}
B.~Neyshabur, R.~R. Salakhutdinov, and N.~Srebro.
\newblock Path-{SGD}: Path-normalized optimization in deep neural networks.
\newblock In \emph{NIPS}, 2015{\natexlab{a}}.

\bibitem[Neyshabur et~al.(2015{\natexlab{b}})Neyshabur, Tomioka, and
  Srebro]{neyshabur2014search}
B.~Neyshabur, R.~Tomioka, and N.~Srebro.
\newblock In search of the real inductive bias: On the role of implicit
  regularization in deep learning.
\newblock In \emph{ICLR}, 2015{\natexlab{b}}.

\bibitem[Neyshabur et~al.(2017)Neyshabur, Tomioka, Salakhutdinov, and
  Srebro]{neyshabur2017geometry}
B.~Neyshabur, R.~Tomioka, R.~Salakhutdinov, and N.~Srebro.
\newblock Geometry of optimization and implicit regularization in deep
  learning.
\newblock \emph{arXiv preprint arXiv:1705.03071}, 2017.

\bibitem[Reinsel and Velu(1998)]{velu2013multivariate}
G.~C. Reinsel and R.~Velu.
\newblock \emph{Multivariate Reduced-Rank Regression: Theory and Applications}.
\newblock Springer Science \& Business Media, 1998.

\bibitem[Saxe et~al.(2013)Saxe, McClellans, and Ganguli]{saxe2013learning}
A.~M. Saxe, J.~L. McClellans, and S.~Ganguli.
\newblock Learning hierarchical categories in deep neural networks.
\newblock In \emph{Proceedings of the Annual Meeting of the Cognitive Science
  Society}, 2013.

\bibitem[Saxe et~al.(2014)Saxe, McClelland, and Ganguli]{saxe2014exact}
A.~M. Saxe, J.~L. McClelland, and S.~Ganguli.
\newblock Exact solutions to the nonlinear dynamics of learning in deep linear
  neural networks.
\newblock In \emph{ICLR}, 2014.

\bibitem[Saxe et~al.(2018)Saxe, McClelland, and Ganguli]{saxe2018mathematical}
A.~M. Saxe, J.~L. McClelland, and S.~Ganguli.
\newblock A mathematical theory of semantic development in deep neural
  networks.
\newblock \emph{arXiv preprint arXiv:1810.10531}, 2018.

\bibitem[Soudry et~al.(2018)Soudry, Hoffer, and Srebro]{soudry2017implicit}
D.~Soudry, E.~Hoffer, and N.~Srebro.
\newblock The implicit bias of gradient descent on separable data.
\newblock In \emph{ICLR}, 2018.

\bibitem[Uschmajew and Vandereycken(2018)]{uschmajew2018critical}
A.~Uschmajew and B.~Vandereycken.
\newblock On critical points of quadratic low-rank matrix optimization
  problems.
\newblock Tech. report (submitted), July 2018.

\bibitem[Zhang et~al.(2017)Zhang, Bengio, Hardt, Recht, and
  Vinyals]{zhang2017understanding}
C.~Zhang, S.~Bengio, M.~Hardt, B.~Recht, and O.~Vinyals.
\newblock Understanding deep learning requires rethinking generalization.
\newblock 2017.

\end{thebibliography}

\newpage
\appendix
\onecolumn

\section{Deep Linear Autoencoder Recovers PCA.} %
\label{sec:deep_linear_autoencoder_recovers_pca_}
Let us recall that the two-layer linear autoencoder can be formulated as,
\begin{equation}\label{eq:MSE_deep_autoencoder}
	(\mW_2^*,\mW_1^*) \in \argmin_{\substack{\mW_2 \in \mathbb{R}^{r \times p}\\ \mW_1 \in \mathbb{R}^{d \times r}}} \frac{1}{2n}  \|\mX -  \mX\mW_1 \mW_2  \|_2^2 \,.
\end{equation}
Thus, the gradients of the objective are,
\begin{equation}	
\nabla f_{\mW_2}(\mW_2,\mW_1)
	=  \mW_1^\top (  \mSigma_x\mW_1\mW_2 - \mSigma_{x})   \quad \text{and} \quad
	\nabla f_{\mW_1}(\mW_2,\mW_1)
	=  (\mSigma_{x}\mW_1\mW_2 - \mSigma_{x}) \mW_2^\top \,.
	\notag
\end{equation}
Thus, if $\mW_2^{(0)} = (\mW_1^{(0)})^\top$ and if $[\mW_1^{(0)}\mW_2^{(0)},\mSigma_x] = 0$, we have that,
\begin{align}
	\nabla f_\mM(\mW_2^{(0)},\mW_1^{(0)})
	&=  (\mW_1^{(0)})^\top (  \mSigma_x\mW_1^{(0)}\mW_2^{(0)} - \mSigma_{x}) \\
	&=  ((\mSigma_{x}\mW_1^{(0)}\mW_2^{(0)} - \mSigma_{x}) (\mW_2^{(0)})^\top )^\top \\
	&= \nabla f_\mN(\mW_2^{(0)},\mW_1^{(0)})^\top \,.
\end{align}
Thus, for the discrete case, by a recurrence we have that, $\mW_1^{(t)} = (\mW_2^{(t)})^\top \,,\;t\geq 0$ and for the continuous case, invoking the Cauchy-Lipschitz theorem, we have that $\mW_1(t) = \mW_2(t)^\top \,,\;t\geq 0$. Consequently, the limit solution is such that 
\begin{equation}\label{eq:MSE_deep_autoencoder}
	\mW_1^*\in \argmin_{ \mW_1 \in \mathbb{R}^{d \times r}} \frac{1}{2n}  \|\mX -  \mX\mW_1 \mW_1^\top  \|_2^2 \,,
\end{equation}
which is a formulation of the PCA.

\section{Proof of Theorems and Propositions} %
\label{sec:proof_of_}
\subsection{Proof of Prop.~\ref{prop:sol_linear_edp}} %
\label{sub:proof_of_proposition_prop:sol_linear_edp}

\begin{repproposition}{prop:sol_linear_edp}
	For any $\mW_0 \in \mathbb{R}^{d \times p}$ , the solution to the linear differential equation~\eqref{eq:edp_linear} is, 
\begin{equation}\label{eq:A_t_non_fact}
	\mW(t) = e^{-t\mSigma_x}(\mW_0-\mSigma_x^\dagger \mSigma_{xy}) + \mSigma_x^{\dagger} \mSigma_{xy} \,,
\end{equation}
where $\mSigma_x^\dagger$ is the pseudoinverse of $\mSigma_x$.
\end{repproposition}

We can differentiate~\eqref{eq:A_t_non_fact} and check if it verifies~\eqref{eq:edp_linear}.
In order to do that, we just need to notice that $\mSigma_x \mSigma_x^{\dagger} \mSigma_{xy} = \mSigma_{xy}$. To see that we compute the SVD of $\mX^\top = \mU^\top \mD \mV$ where $\mD$ is a rectangular matrix with only diagonal coefficients such that,
\begin{equation}
\mD \mD^\top = \diag(\lambda_1,\ldots,\lambda_r,0,\ldots,0)\,. 	
\end{equation}
Thus, we have $\mSigma_x = \mU^\top \diag(\lambda_1,\ldots,\lambda_r,0,\ldots,0)\mU$ and $ \mSigma_x^{\dagger} = \mU^\top \diag(1/\lambda_1,\ldots,1/\lambda_r,0,\ldots,0)\mU$. Leading to,
\begin{equation}\notag
	\mSigma_x \mSigma_x^{\dagger} \mSigma_{xy} = \mU^\top \mD \mV \mY = \mSigma_{xy}\,.
\end{equation}
Consequently, the matrix valued function $\mW(t)$ defined in~\eqref{eq:A_t_non_fact} verifies~\eqref{eq:edp_linear}. Now we just need to use Cauchy-Lipschitz theorem~\citep{coddington1955theory} (a.k.a. Picard–Lindelöf theorem) to say that this solution is the unique solution of the ODE~\eqref{eq:edp_linear}.

\subsection{proof of Thm.~\ref{thm:solution continuous}} %
\label{sub:proof_of_theorem_1}
\paragraph{Commutative case, $\epsilon =0$:}
We use ideas from~\citep{saxe2018mathematical} and combine it with Assum.~\ref{assump:commutativity} for $\epsilon = 0$. Note that $\epsilon =0$ if and only if $\mSigma_x$ and $\mSigma_{xy}$ commute. 
thus, we have that, 
\begin{equation}\label{eq:proof_basis}
	\mSigma_{xy} = \mU \mD_{xy} \mV^\top
	\quad \text{and} \quad 
	\mSigma_{x} = \mU \mD_x \mU^\top \,.
\end{equation}
Let us consider a generalization of the linear transformation proposed by~\citet[Eq. S6,S7]{saxe2018mathematical},
\begin{equation}\label{eq:app_linear_transformation_M_N}
 	\bar \mW_1 = \mU^\top \mW_1 \mQ_1 \,, \quad
 	\bar \mW_l = \mQ_{l-1}^{-1} \mW_l \mQ_{l} , \; 2\leq l \leq L-1 \,,
 	\quad \text{and} \quad
 	\mW_L = \mQ_{L-1}^{-1} \mW_L \mV \,,
\end{equation}
where $\mQ_l \, ,\; 1 \leq l \leq L-1$ are arbitrary invertible matrices. Then, noting $\mQ_0 := \mU$ and $\mQ_L := \mV$, we get the following dynamics,
\begin{equation}
	 \frac{d\bar \mW_l(t)}{dt} = \mQ_{l-1}^{-1}\mW_{1:l-1}(t)^\top \! (\mSigma_{xy} \! - \! \mSigma_x\mW(t)) \mW_{l+1:L}(t)^\top  \mQ_l
	\,, \quad 1\leq l \leq L\,.
\end{equation}
Thus using~\eqref{eq:proof_basis}, the fact that $\mU^\top \mU = \mI_d$ and that for any invertible matrix $\mQ$, we have $(\mQ^{-1})^\top = (\mQ^\top)^{-1}$, we get that,
\begin{equation}\label{eq:edp_deep_change}
	 \frac{d\bar \mW_l(t)}{dt} = \bar\mW_{1:l-1}(t)^\top \big(\mD_{xy} -  \mD_x \bar\mW(t)\big) \bar \mW_{l+1:L}(t)^\top \,, \quad  \mW_l(0) = \mW_l^{(0)}
	 \;1\leq l\leq L \,.
\end{equation}
Using the same argument as~\citep{saxe2018mathematical}, if $\bar \mW_l(t)\,,\;1\leq l\leq L$ only have diagonal coefficients then their derivative also only have diagonal coefficients. Thus, if we initialize $\mW_l^{(0)} \,,\;1\leq l\leq L$, only with diagonal coefficients we have a decoupled solution for each diagonal coefficient. This argument can be formalized using Cauchy-Lipschitz theorem: \eqref{eq:edp_deep_change} has a unique solution which is the one we will exhibit in the following.

Recall that we noted $r_0 = d$ and $r_L = p$ and that $\mW_l \in \mathbb{R}^{r \times {l-1},r_l}$.
Let us note, $r =\min \{r_l \,:\, 0\leq l\leq L-1\}$ and $w_{l,i}(t), \, 1\leq i \leq r$ the respective diagonal coefficients of $\mW_l(t)$ for $1\leq l\leq L$. Note that for $i \geq r$ one could define diagonal coefficients for some of the matrices $\mW_l$ but their gradient will be equal to $0$, thus non-trivial dynamics only occur for $i\leq r$. 
 They follow the equation, 
\begin{equation} \label{eq:edp_coeffs}
	\dot w_{l,i}(t) = w_{-l,i}(t)(\sigma_i - \lambda_i w_i(t))\,, \; w_{l,i}(0) \in \R \,, \quad 1\leq l \leq L\,, \quad 1\leq i \leq r \,,
\end{equation}
where the notation $w_{-l,i}(t)$ stands for the product of the $w_{k,i}(t) \,,\, 1\leq k\leq L$ omitting $w_{l,i}(t)$, i.e., 
\begin{equation}
w_{-l,i}(t) := \prod_{\substack{k=1\\k\neq l}}^L w_{k,i}(t) \,,
\end{equation}
and $w_{i}(t)$ stands for the product of the $w_{k,i}(t)\,,\, 1\leq k\leq L$.
The difference with \citep{saxe2018mathematical} is that, since they only consider the case $\mSigma_x = \mI_d$ they have $\lambda_i = 1$, they also only consider the case $L=2$. The use of Assumption~\ref{assump:commutativity} allowed us to work in a more general case.

We will assume that if $w_{l,i}(t) = w_{k,i}(t) \,, \, 1\leq l,k \leq L$, to find an analytic solution and then show that if $w_{l,i}(0) = w_{k,i}(0) \,,\, 1\leq k,l\leq L$ then this analytic solution verifies~\eqref{eq:edp_coeffs} and thus, by Cauchy-Lipschitz theorem, is the unique solution of the non-linear differential equation.

Thus, considering $w_i(t) := w_{1,i}(t) \cdots w_{L,i}(t)$, and assuming that $w_{l,i}(t) = w_{k,i}(t) \,, \, 1\leq l,k \leq L$, we get that, for $1\leq i \leq r,$ 
\begin{align} \label{eq:w_differential}
	\dot w_i(t) 
	&= \sum_{l=1}^L w_{1,i}(t) \cdots w_{l-1,i}(t) \dot w_{l,i}(t) w_{l+1,i}(t) \cdots w_{L,i}(t)  \\
	&= L w_i(t)^{2-2/L} (\sigma_i - \lambda_i w_i(t)) \,,\;w_i(0) \in \R  \,.
\end{align}
\begin{lemma} If $w_i(0) \in (0,\frac{\sigma_i}{\lambda_i})$, then
the differential equations~\label{eq:w_differential} has a unique solution that is increasing and $w_i(t) \in (0,\frac{\sigma_i}{\lambda_i}) \,,\, \forall t \in \R$.
\end{lemma}
\proof If at a time $t \in \R$, we have $w_i(t) = 0$ and thus $\dot w_i(t)$. Noticing that then the constant function $w_i(t) = 0 \, t\in \R$ is a solution of~\eqref{eq:w_differential}, by Cauchy-Lipschitz it is the only one. We can use the same argument to say that if there exists a time $t \in \R$, such we have $w_i(t) = 0$ then $w_i(t) = 0 \, \forall t\in \R$. Thus by continuity of $w_i(t)$ we have that if $w_i(0) \in (0,\frac{\sigma_i}{\lambda_i})$ then, $w_i(t) \in (0,\frac{\sigma_i}{\lambda_i}) \,,\, \forall t \in \R$ .
\endproof

\paragraph{Case $L=2$:} in that case we have two situations, $\sigma_i > 0$ and $\sigma_i = 0 \,,\; \lambda_i > 0$ (the case $\sigma_i = \lambda_i =0$ give a constant functions). 

For $\sigma_i >0$ we have that,
\begin{align}
	t 
	&= \int_{0}^t \frac{d w_i(t) }{2 w_i(t) (\sigma_i - \lambda_i w_i(t))}\\
	&=  \frac{1}{2 \sigma_i}\int_{0}^t \frac{ dw_i(t }{ w_i(t) } +  \frac{\lambda_i dw_i(t}{\sigma_i - \lambda_i w_i(t)}\\
	&= \frac{1}{2 \sigma_i} \ln \frac{w_i(t)(\sigma_i - \lambda_i w_i(0) )}{w_i(0) (\sigma_i - \lambda_i w_i(t) )} \,.
\end{align}
Leading to, 
\begin{equation}
	w_i(t) = \frac{w_i(0) \sigma_i e^{2\sigma_i t}}{ w_i(0) \lambda_i (e^{2\sigma_i t} -1)+ \sigma_i} \,.
\end{equation}
In order to get a solution for $w_{2,i}(t)$ and $w_{1,i}(t)$, we will use Cauchy-Lipschitz theorem~\citep{coddington1955theory}. The idea is that if we find a solution of~\eqref{eq:edp_coeffs}, it is the only one. Let us set, $m_i(0) = n_i(0)= e^{-\delta_i}$, then we can set,
\begin{equation}
	w_{2,i}(t) =  w_{1,i}(t) = \sqrt{\frac{ \sigma_i e^{2\sigma_i t - 2 \delta_i}}{ \lambda_i (e^{2\sigma_i t - 2 \delta_i} - e^{-2 \delta_i})+ \sigma_i}} \,,
\end{equation}
and verify that we have,
\begin{align} 
	&\dot w_{2,i}(t) = w_{1,i}(t)(\sigma_i - \lambda_i w_{1,i}(t) w_{2,i}(t)) \,, \; m_i(0) = e^{-\delta_i}\\
	&\dot w_{1,i}(t) = w_{2,i}(t)(\sigma_i - \lambda_i w_{1,i}(t) w_{2,i}(t))  \,, \; n_i(0) = e^{-\delta_i} \quad 1\leq i \leq r \,.
\end{align}
Thus, this is the unique solution of~\eqref{eq:edp_coeffs}.

For $\sigma_i =0\,,\; \lambda_i >0$ we have that,
\begin{equation}
	t = \int_{0}^t \frac{\dot w_i(t) }{- 2\lambda_i w_i(t)^2} dt
	= \frac{1}{ 2 \lambda_i}\left(\frac{1}{w_i(t)} -  \frac{1}{w_i(0)}\right)\,.
\end{equation}
Thus,
\begin{equation}
	w_i(t) = \frac{w_i(0)}{1 + 2 w_i(0) \lambda_i t} \,.
\end{equation}
Thus, if we initialize with $m_i(0) = n_i(0) = e^{-\delta_i}$ we get,
\begin{equation}
	w_{1,i}(t) = w_{2,i}(t) = \frac{e^{-\delta_i}}{\sqrt{1 + 2 e^{-\delta_i} \lambda_i t}}\,.
\end{equation}

\paragraph{Non commutative case $\epsilon >0$.} Now, we will consider Assumption~\ref{assump:commutativity} with $\epsilon >0$ and $L=2$. 

First let us proove two lemmas usefull for later,
\begin{lemma}The matrix valued function $\mW(t)$ converge to $\mX^\dagger\mY$ and thus is bounded for $t>0$.
\end{lemma}
\proof
Since~\eqref{eq:deep_continuous_dynamics} is a \emph{gradient dynamics}, it only moves in the span of the gradient of $f$ (the explicit expressions of $\nabla f$ is derived in~\eqref{eq:grad_g} for $L=1$ and~\eqref{eq:grad_f} for a general $L$). We use this property to characterize the solution found by these dynamics. We can study each column of the predictors $\mW := \mW_1 \cdots \mW_L$.
If we look at the columns of $\nabla_{\mW_L} f$, they are included in $\mX^\top$, thus it means that if we initialize the columns of $\mW_L^{(0)}$ in that span, then the columns of $\mW$ will belong to that span during the whole learning process, 
\begin{equation}
\label{eq:stay_in_span}
\begin{aligned}
&[\mW]_i \in \vecspan(\nabla_{\mW_L} f) \subset \vecspan(\mX^\top), \; 1\leq i \leq n\,,
\end{aligned}
\end{equation}
where $\mW$ is $\mW(t)$.
Thus, if the dynamics~\eqref{eq:deep_continuous_dynamics} converge, then they converge to a matrix with the $i^{th}$ column vector being in the intersection,
\begin{equation}\label{eq:finds_min_norm_solution}
\vecspan(\mX^\top) \cap \{\vu : \mX \vu = [\mY]_i\}	=\{ \mX^\dagger [\mY]_i \} \,.
\end{equation}
Finally, we have $\mX\mW(t) \to \mY$ by definition of the gradient dynamics,
\begin{equation}
	\frac{d \|\mY- \mX \mW_1(t) \mW_2(t)\|^2}{dt} =  - \|\nabla_{\mW_1} f(\mW_1(t),\mW_2(t))\|^2  - \|\nabla_{\mW_2} f(\mW_1(t),\mW_2(t))\|^2<0
\end{equation}
\endproof

\begin{lemma}\label{lemma:compact}
We have that $\| \mW_1(t)\|^2 = O(t)$ and $\| \mW_2(t)\|^2 = O(t)$.
\end{lemma}
\proof We have that,
\begin{equation}
  \frac{d \| \mW_1(t)\|^2}{dt} 
   = \langle \mW_1(t), (\mSigma_{xy} -  \mSigma_x \mW(t))\mW_2(t)^\top \rangle
   = Tr((\mSigma_{xy} -  \mSigma_x \mW(t))\mW(t)^\top)\,.
\end{equation}
Since $\mW(t)$ is bounded then $\| \mW_1(t)\|^2 = O(t)$. The same way we have $\| \mW_2(t)\|^2 = O(t)$
\endproof

After the same change of basis as in the commutative case The matrices $\bar \mW_l(t)$ follow the differencial equations
\begin{equation}\label{eq:edp_proof}
	 \frac{d\bar \mW_1(t)}{dt} =  \big(\mD_{xy} -  (\mD_x + \mB) \bar\mW(t)\big) \bar \mW_{2}(t)^\top , \quad
	 \frac{d\bar \mW_2(t)}{dt} = \bar \mW_{1}(t)^\top  \big(\mD_{xy} -  (\mD_x + \mB) \bar\mW(t)\big)
\end{equation}

In order to perform pertrubation analysis we will use a consequence of Grönwall's inequality~\citep{gronwall1919note}.
\begin{lemma} 
\label{lemma:gronwall}
Let $\beta$ be a non negative function and $\alpha$ a non decreasing function. Let $u$ be a function defined on an interval $I = [a,\infty)$ such that 
\begin{equation}
  u(t) \leq \alpha(t) + \int_a^t \beta(s) u(s) ds \,,\quad \forall t \in I\,.
\end{equation}
then we have that 
\begin{equation}
 	u(t) \leq \alpha(t)\exp\left(\int_a^t \beta(s) ds\right)\,, \quad \forall t \in I\,.
 \end{equation} 
\end{lemma}
\proof The proof can be found for instance in \citep[Lemma 3.1.6]{berglund2001perturbation} \endproof
\endproof

Thus, let us consider $\bar \mW_1(t)$ and $\bar \mW_2(t)$ the solutions of~\eqref{eq:edp_proof} as well as $\bar \mW^0_1(t)$ and $\bar \mW^0_2(t)$ the solutions of the very same differential equation but with $\mB =0$. For notational simplicity we will omit the bar on the matrices $\mW$ in the following. We have that,
\begin{equation}
	 \mW_1(t) -  \mW^0_1(t) 
	= \int_0^t [-\mB  \mW(s)  \mW_2(s) + (\mD_{xy} -  \mD_x  \mW(s))  \mW_2(s)^\top - (\mD_{xy} -  \mD_x  \mW^0(s))  \mW_2^0(s)^\top ] ds \notag
\end{equation}
Leading to
\begin{align*}
	\| \mW_1(t) -  \mW^0_1(t)\| 
	&\leq \int_0^t  \|D_x\| \|\mW_1(s) \mW_2(s) \mW_2(s)^\top - \mW_1^0(s) \mW_2^0(s) \mW_2^0(s)^\top\| ds\\
	& \quad + \int_0^t \|D_xy\|\|\mW_2(s)-\mW_2^0(t)\| ds + \|\mB  \mW(t)  \mW_2(t)\| 
\end{align*}
In order to upper bound the first integral we will consider the function $F(\mA,\mB):= \mA \mB \mB^\top$. This function is Lipschitz on any compact because this function is infinitely differentiable. Thus we have that (omiting the $t$ in the notation),
\begin{align}
\|\mW_1 \mW_2 \mW_2^\top - \mW_1^0 \mW_2^0 (\mW_2^0)^\top\| 
&= t^{3/2}\|\tfrac{\mW_1}{\sqrt{t}} \tfrac{\mW_2}{\sqrt{t}} \tfrac{\mW_2}{\sqrt{t}}^\top - \tfrac{\mW_1^0}{\sqrt{t}} \tfrac{\mW_2^0}{\sqrt{t}} (\tfrac{\mW_2^0}{\sqrt{t}})^\top\| \\
&= t^{3/2}\|F(\frac{ \mW_1}{\sqrt{t}},\frac{ \mW_2}{\sqrt{t}}) - F(\frac{ \mW_1^0}{\sqrt{t}},\frac{ \mW_2^0}{\sqrt{t}})\| \notag\\
&\leq t^{3/2}L(\|\tfrac{\mW_2}{\sqrt{t}}-\tfrac{\mW_2^0}{\sqrt{t}}\| + \|\tfrac{\mW_1}{\sqrt{t}}-\tfrac{\mW_1^0}{\sqrt{t}}\|) \notag\\
& \leq t L(\|\mW_2-\mW_2^0\| + \|\mW_1-\mW_1^0\|) \notag
\end{align}
Using the fact that $\frac{ \mW_1(t)}{\sqrt{t}}, \frac{ \mW_1^0(t)}{\sqrt{t}},\frac{ \mW_2^0(t)}{\sqrt{t}}$ and $\frac{ \mW_2(t)}{\sqrt{t}}, \, t\geq 0$ live in a compact set (Lemma~\ref{lemma:compact}) and that $F$ is Lipschitz on any compact.
Thus, using that $\|S\| = O(\epsilon)$, we have
\begin{equation}
\| \mW_1(t) -  \mW^0_1(t)\| 
	\leq O(\epsilon) + O(1) \int_0^t  s (\|\mW_2(s)-\mW_2^0(s)\| + \|\mW_1(s)-\mW_1(s)^0\|)ds \notag
\end{equation}
The same way we can prove that, 
\begin{align*}
	\| \mW_2(t) -  \mW^0_2(t)\| 
	\leq O(\epsilon) + O(1) \int_0^t  s (\|\mW_2(t)-\mW_2^0(t)\| + \|\mW_1(t)-\mW_1^0(t)\|)dt
\end{align*}
And consequently we can sum these two inequalities and apply Grönwall's inequality with the quantities $u(t)= \| \mW_1(t) -  \mW^0_1(t)\| + \| \mW_2(t) -  \mW^0_2(t)\| $, $\alpha(t) = O(\epsilon)$ and $\beta(s) = O(s)$ to get,
\begin{equation}
\| \mW_1(t) -  \mW^0_1(t)\| + \| \mW_2(t) -  \mW^0_2(t)\| 
	\leq \epsilon \cdot e^{O(t^{2})}
\end{equation}

\subsection{Proof of Thm.~\ref{thm:eig_values}} %
\label{sub:proof_of_theorem_eig_vlaues}

\begin{reptheorem}{thm:eig_values} Let us denotes $w_i(t)$, the values defined in~\eqref{eq:m_i_ni}. If $m_i(0) = e^{-\delta}\,,\;1\leq i\leq r,$ then we have,
\begin{equation}
	w_i(\delta t) \underset{\delta \to \infty}{\to} \left\{ 
	\begin{aligned} 0 & \quad \text{if} \quad t < 1/\sigma_i   \\
	\sqrt{\tfrac{\sigma_i}{\lambda_i + \sigma_i}} & \quad \text{if} \quad t = 1/\sigma_i \\
	\sqrt{\tfrac{\sigma_i}{\lambda_i}} & \quad \text{otherwise} \,.
	\end{aligned}
	\right. \notag
\end{equation}
\end{reptheorem}
\proof Using~\eqref{eq:m_i_ni} we get that,
\begin{equation}
	w_i(\delta t) = \sqrt{\frac{ \sigma_i e^{2\delta(\sigma_i t - 1)}}{ \lambda_i (e^{2 \delta(\sigma_i t - 1)} - e^{-2 \delta})+ \sigma_i}} \,.
\end{equation}
Then we can conclude saying that for any $i$ and $t\geq0$,
\begin{equation}
	e^{2\delta(\sigma_i t - 1)}
	\underset{\delta \to \infty}{\to} \left\{ \begin{aligned} 0 & \quad \text{if} \quad  t < {1}/{\sigma_i}  \\
	1 & \quad \text{if} \quad t = {1}/{\sigma_i}  \\
	+\infty & \quad \text{otherwise} \,,
	\end{aligned}
	\right. 
\end{equation}
and that when $\delta \to \infty$,
\begin{equation}
\|\mW_i^0(\delta t) - \mW_i^\epsilon(\delta t)\|
 \leq \epsilon \cdot  e^{c t^{2}} = e^{\delta^{2}(c t^{2} - \ln(\delta))} \to 0
\end{equation}

\endproof
\subsection{Proof of Eq.~\ref{eq:result_explicit}} %
\label{sub:proof_of_eq_eq:result_explicit}

Let us recall~\eqref{eq:result_explicit},
\begin{equation}\tag{\ref{eq:result_explicit}}
	\mW_t= (\mW_0-\mSigma_x^\dagger \mSigma_{xy})(\mI_d-\eta \mSigma_{x})^{t} +\mSigma_x^\dagger\mSigma_{xy}\,.
\end{equation}
Thus we have that,
\begin{align}
	\mW_t 
	&= \mW_0 (\mI_d-\eta \mSigma_{x})^{t} +  \eta \mSigma_{xy} \sum_{s=0}^{t-1} (\mI_d-\eta \mSigma_{x})^{s} \\
	&= \mW_0 (\mI_d-\eta \mSigma_{x})^{t} +  \mSigma_x^\dagger \mSigma_{xy}  -  \mSigma_x^\dagger \mSigma_{xy} (\mI_d-\eta\mSigma_{x})^t \notag  \\
	&= (\mW_0-\mSigma_x^\dagger \mSigma_{xy})(\mI_d-\eta \mSigma_{x})^{t} +\mSigma_x^\dagger\mSigma_{xy}\,.
\end{align}

\subsection{Proof of Thm.~\ref{thm:discrete_case}} %
\label{sub:proof_of_theorem_3}

\paragraph{Case $\epsilon=0$.}
If we define $\mW^{(t)} := \mW_1^{(t)}\mW_2^{(t)}$, the discrete update scheme for the \emph{two-layer linear neural network}~\eqref{eq:MSE_deep} is,
\begin{equation} \label{eq:updates_discrete_appendix}
\left\{
\begin{aligned}
	&\mW_1^{(t+1)} = \mW_1^{(t)}\! - \eta  (\mSigma_{x}\mW^{(t)}\! - \mSigma_{xy}) (\mW_2^{(t)})^\top \\
	&\mW_2^{(t+1)} = \mW_2^{(t)}\! - \eta  (\mW_1^{(t)})^\top (  \mSigma_x\mW^{(t)} \! - \mSigma_{xy}) \, .  
\end{aligned}
\right.
\end{equation}
Using the same transformation~\eqref{eq:app_linear_transformation_M_N} as in \S\ref{sub:proof_of_theorem_1} we get that,
\begin{equation}
\left\{
\begin{aligned}
	&\bar \mW_1^{(t+ 1)} = \bar \mW_1^{(t)}\! - \eta  (\mD\bar \mW^{(t)}\! - \mS) (\bar \mW_2^{(t)})^\top \\
	&\bar \mW_2^{(t+1)} = \bar \mW_2^{(t)}\! - \eta  (\bar \mW_1^{(t)})^\top (  \mD\bar \mW^{(t)} \! - \mS) \, .  
\end{aligned}
\right.
\end{equation}
where $\mD$ and $\mS$ only have diagonal coefficients. Thus, by an immediate recurrence we can show that if $\bar \mW_1^{(0)}$ and $\bar \mW_2^{(0)}$ only have diagonal coefficients then $\bar \mW_1^{(t)}$ and $\bar \mW_2^{(t)}$ for $t \in \N$ only have diagonal coefficients.

let us note, $r = \min(p,d)$ and $m_1^{(t)},\ldots,m_r^{(t)}$ and $n_1^{(t)},\ldots,n_r^{(t)}$ the respective diagonal coefficients of $\bar \mW_1^{(t)}$ and $\bar \mW_2^{(t)}$, they follow the equation, 
\begin{equation} \label{eq:discrete_rec_coeffs}
	m_i^{(t+1)} = m_i^{(t)} + \eta n_i^{(t)}(\sigma_i - \lambda_i n_i^{(t)} m_i^{(t)}) 
	\quad \text{and} \quad 
	n_i^{(t+1)} = n_i^{(t)} + \eta m_i^{(t)}(\sigma_i - \lambda_i n_i^{(t)} m_i^{(t)})  \, , \quad 1\leq i \leq r \,.
\end{equation}

In order to prove Thm.~\ref{thm:discrete_case}, we will prove several properties on the sequences $(m_i^{(t)})_{t\geq 0}$ and $(n_i^{(t)})_{t\geq 0}\,,\;1\leq i\leq d$.  First let us introduce the sequence $(a_i^{(t)})_{t\geq 0}$ defined as $a_i^{(t)} := m_i^{(t)} n_i^{(t)} \,, t \geq 0$.
\begin{lemma}\label{lemma:equal} If $m_i^{(0)} = n_i^{(0)}$ then,
\begin{equation}
	m_i^{(t)} = n_i^{(t)} = \sqrt{a_i^{(t)}} \, \qquad \forall t \in \N \,.
\end{equation}
\end{lemma}	
\proof By a straightforward recurrence we have that if at time $t$, $m_i^{(t)} = n_i^{(t)}$ then by~\eqref{eq:discrete_rec_coeffs}, we have $m_i^{(t+1)} = n_i^{(t+1)}$.
\endproof
 Thus, we will now focus on the sequence $(a_i^{(t)})_{t\geq 0}$, by~\eqref{eq:discrete_rec_coeffs}, we have that
 \begin{equation}\label{eq:rec_a_i_t}
 	a_i^{(t+1)} 
 	= a_i^{(t)} + 2\eta a_i^{(t)}(\sigma_i - \lambda_i a_i^{(t)}) + \eta^2 a_i^{(t)}(\sigma_i - \lambda_i a_i^{(t)})^2
 	=  a_i^{(t)} + \eta a_i^{(t)}(\sigma_i - \lambda_i a_i^{(t)})(2 + \eta(\sigma_i - \lambda_i a_i^{(t)}))\,.
 \end{equation}

Similarly as for the continuous case, there is two different behavior $\sigma_i >0$ ad $\sigma_i =0\,, \; \lambda_i >0$. In the following we assume that $\eta >0$.

For $\sigma_i >0$ we can derive the following results,
\begin{lemma}
For any $1\leq i \leq r_{xy}$, if $0<a_i^{(0)} < \frac{\sigma_i}{\lambda_i}$ and $2\eta \sigma_i<1$, then, the sequence $(a_i^{(t)})$ is increasing and 
\begin{equation}\label{eq:mu_bounded}
	0 < a_i^{(t)} < \frac{\sigma_i}{\lambda_i} \, ,\quad \forall t \geq 0 \,.
\end{equation}
\end{lemma}
\proof By assumption~\eqref{eq:mu_bounded} is true for $t=0$. 

Let us assume that~\eqref{eq:mu_bounded} is true for a time-step $t$ and let us prove that it is still true at time-step $t+1$.

Using the recursive definition~\eqref{eq:rec_a_i_t} of $a_i^{(t)}$ we get for $t \geq 0$,
\begin{align}
	a_{i}^{(t+1)} 
	&= a_i^{(t)} + \eta a_i^{(t)} (\sigma_i- \lambda_ia_i^{(t)})(2+ \eta (\sigma_i- \lambda_ia^{(t)}_i)) \\
	&>  a_i^{(t)} > 0 \,.
\end{align}
For the upper bound we need to notice that $a_{i}^{(t+1)} = f_i(a_{i}^{(t)})$ where $f_i: x \mapsto x + \eta x(\sigma_i- \lambda_i x)(2 + \eta (\sigma_i- \lambda_i x))$ where $\eta  >0$. Since we assumed that $2\eta \sigma_i<1$, we have that,
\begin{align}
	f_i(x) 
	&< x + \eta  x(\sigma_i- \lambda_i x)(2 + \tfrac{\sigma_i- \lambda_ix}{2 \sigma_i}) \, , \quad \forall x \in (0,1) \\
	& < x +  \frac{x(1- \frac{\lambda_i}{\sigma_i}x)(\frac{5}{2} - \frac{\lambda_i}{2\sigma_i}x)}{2} =: g_i(x) \, , \quad \forall x \in (0,1) \,.\\
\end{align}
Then we just need to show that $g(x) <\tfrac{\lambda_i}{\sigma_i}\,,\; \forall x\in(0,\tfrac{\lambda_i}{\sigma_i})$.
\begin{equation}
	 4g'(x) = 9 - 12\tfrac{\lambda_i}{\sigma_i}x + 3\tfrac{\lambda_i^2}{\sigma_i^2}x^2 > 0 \, ,\; \forall x \in (0,\tfrac{\lambda_i}{\sigma_i}) \,.
\end{equation}
Thus $g$ is  non-decreasing on $(0,1)$ and consequently, $g(x)<g(\tfrac{\lambda_i}{\sigma_i}) = \tfrac{\lambda_i}{\sigma_i}\,,\; \forall x \in (0,1)$.

Finally, we get that, 
\begin{equation}
	a_{i}^{(t+1)} = f_i(a_{i}^{(t)}) < g(a_{i}^{(t)}) < g(\tfrac{\lambda_i}{\sigma_i}) = \frac{\lambda_i}{\sigma_i} \,.
\end{equation}
\endproof

With this lemma we can proof Thm.~\ref{thm:discrete_case}. Let us first recall this theorem.

\begin{reptheorem}{thm:discrete_case}
	For any $1\leq i \leq r_{xy}$, if $ \frac{\sigma_i}{\lambda_i} >a_i^{(0)} >0$ and $2\eta \sigma_i<1$, then $ \forall t \geq 0 \,,\; 1 \leq i \leq r $ we have,
\begin{align}\label{eq:mu_asym_upperbound}
	 &a_i^{(t)} \leq  \frac{a_i^{(0)}}{(\sigma_i- \lambda_i a_i^{(0)})e^{(-2 \eta \sigma_i  -  \eta^2 \sigma_i^2 )t } + a_i^{(0)} \lambda_i}\\
	\text{and} \qquad
	&a_i^{(t)} \geq  \frac{a_i^{(0)}}{(\sigma_i- \lambda_i a_i^{(0)})e^{(-2 \eta \sigma_i  + 4 \eta^2 \sigma_i^2 )t } + a_i^{(0)} \lambda_i}\,,
	\label{eq:mu_asym_lowerbound}
\end{align}
and for $r_{xy} \leq i \leq r_x$,
\begin{align*}\label{eq:mu_sublinear}
	&a_i^{(t)} \leq  \frac{a_i^{(0)}}{1 + a_i^{(0)} \lambda_i \eta t} \,.
\end{align*}
\end{reptheorem}
\proof 
In this proof for notational compactness we will remove the index $i$.

We first prove~\eqref{eq:mu_asym_upperbound}, we work with $1/a^{(t+1)}-\frac{\lambda}{\sigma}$, Using~\eqref{eq:rec_a_i_t} we get,
\begin{align}
	1/a^{(t+1)}- \frac{\lambda}{\sigma}	  
	& = \frac{1}{a^{(t)}} \left( \frac{1}{1 + 2 \eta \sigma  (1- \frac{\lambda}{\sigma} a^{(t)}) + \eta^2 \sigma^2 (1- \frac{\lambda}{\sigma} a^{(t)})^2} \right)   - \frac{\lambda}{\sigma}	\\
	& \geq \frac{1}{a^{(t)}} \left( \frac{1}{1 + (2 \eta \sigma + \eta^2 \sigma^2) (1- \frac{\lambda}{\sigma} a^{(t)})} \right) - \frac{\lambda}{\sigma}	\\
	& \geq \frac{1}{a^{(t)}} - \frac{\lambda}{\sigma}  -  \frac{2 \eta \sigma + \eta^2 \sigma^2}{a^{(t)}} (1- \frac{\lambda}{\sigma} a^{(t)})	\,,
\end{align}
where we used that $\frac{1}{1+x} \geq 1- x \,,\; \forall x \geq 0$. Thus we have,
\begin{align}
	1/a^{(t)}- \frac{\lambda}{\sigma}	  
	& \geq (\frac{1}{a^{(t-1)}} - \frac{\lambda}{\sigma} )(1- 2 \eta \sigma - \eta^2 \sigma^2) \\
	& \geq (\frac{1}{a^{(0)}} - \frac{\lambda}{\sigma} )(1- 2 \eta \sigma - \eta^2 \sigma^2)^{t} \\
	& \geq (\frac{1}{a^{(0)}} - \frac{\lambda}{\sigma} )e^{(- 2 \eta \sigma - \eta^2 \sigma^2)t}\,.
\end{align}
Thus Leads to,
\begin{equation}
	a^{(t)} \leq \frac{\sigma a^{(0)}}{(\sigma-{\lambda}a^{(0)}) e^{t(- 2 \eta \sigma -  \eta^2 \sigma^2)} + a^{(0)}\lambda} \;.
\end{equation}
To prove~\eqref{eq:mu_asym_upperbound} we will once again work with $1/a^{(t+1)}- \frac{\lambda}{\sigma}$. Using~\eqref{eq:rec_a_i_t} we get 
\begin{align}
	1/a^{(t+1)}-\frac{\lambda}{\sigma}	  
	& = \frac{1}{a^{(t)}} \left( \frac{1}{1 + 2 \sigma (1- \frac{\lambda}{\sigma} a^{(t)}) + \sigma^2 (1-\frac{\lambda}{\sigma}a^{(t)})^2} \right)   -\frac{\lambda}{\sigma}		\\
	& \leq \frac{1}{a^{(t)}} \left( \frac{1}{1 + 2 \sigma (1- \frac{\lambda}{\sigma} a^{(t)})} \right) - \frac{\lambda}{\sigma}	 \\
	& \leq \frac{1}{a^{(t)}} - \frac{\lambda}{\sigma}	  -  \frac{2 \eta \sigma}{a^{(t)}} (1- \tfrac{\lambda}{\sigma} a^{(t)})	+ \frac{4\eta^2 \sigma^2}{a^{(t)}} (1- \tfrac{\lambda}{\sigma} a^{(t)})^2 \,,
\end{align}
where we used that $\frac{1}{1+x} \leq 1- x + x^2 \,,\; \forall x \geq 0$. Thus we have,
\begin{align}
	1/a^{(t)}-\frac{\lambda}{\sigma}  
	& \leq (\frac{1}{a^{(t)}} - \frac{\lambda}{\sigma} )(1- 2 \sigma \eta + 4 \eta^2\sigma^2) \\
	& \leq (\frac{1}{a^{(0)}} - \frac{\lambda}{\sigma} )(1- 2 \sigma \eta + 4 \eta^2\sigma^2)^{t} \\
	& \leq (\frac{1}{a^{(0)}} - \frac{\lambda}{\sigma} )e^{(- 2 \sigma \eta + 4 \eta^2\sigma^2)t}\,.
\end{align}
This leads to,
\begin{equation}
	a^{(t)} \geq \frac{\sigma a^{(0)}}{(\sigma- \lambda a^{(0)}) e^{t(- 2 \sigma + 4 \eta^2\sigma^2)} +\lambda a^{(0)}} \,.
\end{equation}

Now for $\sigma= 0$ and $\lambda>0$ we have that,
\begin{equation}
	a_i^{(t+1)} 
 	= a_i^{(t)}(1 - 2 \lambda \eta a_i^{(t)}+ \lambda^2 \eta^2 (a_i^{(t)})^2)  \,.
\end{equation}
Thus, considering $(a_i^{(t)})^{-1}$ we get
\begin{align}
	1/a_i^{(t+1)} 
 	&= 1/a_i^{(t)}(1 - 2 \lambda \eta a_i^{(t)}+ \lambda^2 \eta^2 (a_i^{(t)})^2)^{-1} \\
 	&\geq  1/a_i^{(t)}(1 +  2 \lambda \eta a_i^{(t)}- \lambda^2 \eta^2 (a_i^{(t)})^2) \\
 	& = 1/a_i^{(t)} + 2 \lambda \eta - \lambda^2 \eta^2 a_i^{(t)} \,.
\end{align}
Thus, if we assume that $1/a_i^{(0)} \geq \lambda \eta$ we have that $(1/a_i^{(t)})$ is a increasing sequence and that,
\begin{equation}
	1/a_i^{(t)} \geq  1/a_i^{(t-1)} + \lambda \eta  \geq 1/a_i^{(0)} + \lambda \eta t \,,
\end{equation}
leading to,
\begin{equation}
	a_i^{(t)} \leq \frac{a_i^{(0)}}{1+ a_i^{(0)} \lambda \eta t} \,.
\end{equation}
\endproof 

\paragraph{Case $\epsilon >0$.} If we are able to show that all the sequences $\mW_1^{(t)}$ and $\mW_2^{(t)}$ are bounded 

From this theorem we can deduce the following corollary,

\paragraph{Proof of Corollary~\ref{cor:discrete_case}}
Let us recall Corollary~\ref{cor:discrete_case}.
\begin{repcorollary}{cor:discrete_case}
If $\eta < \frac{1}{2\sigma_1}$, $\eta < 2\frac{\sigma_{i}- \sigma_{i+1}}{\sigma_i^2}$ and $\eta < \frac{\sigma_{i}- \sigma_{i+1}}{2\sigma_{i+1}^2} \,,\; \forall i \;1\leq i \leq r_{xy} -1$ then for $1 \leq i < r_{x}$, 
\begin{equation}
	a_{i}^{(\delta T_j)} \to \underset{\delta \to \infty}{\to} \left\{ \begin{aligned} &0  \quad \text{if} \quad  i > r_{xy}\quad \text{or} \quad  j <i  \\
	&\frac{\sigma_i}{\lambda_i}  \quad \text{if} \quad  i \leq r_{xy} \; \,\; \text{and} \;\;\,  j>i \,,
	\end{aligned}
	\right. \notag
\end{equation}
where $T_j := \frac{1}{\sigma_j \eta} \, ,\; 1 \leq j \leq r_{xy}$ and $T_j = + \infty \text{ if } j > r_{xy}$ and $T_0=0$.
\end{repcorollary}
\proof First let us notice that since $\sigma_1 > \ldots > \sigma_{r_{xy}} >0$, the assumption $\eta < 1 /(2\sigma_1)$ implies $\eta < 1 /(2\sigma_i) \,, \; 1\leq i \leq d$. 

Let $i \leq r_{xy}$.
Let us first prove that if  $j<i$, then $a_{i}^{(T_j)}  \underset{\delta \to \infty}{\to} 0$.

Using~\eqref{eq:mu_asym_upperbound} and recalling that in Thm.~\ref{thm:discrete_case}, we assume that $a_i^{(0)} = e^{-2\delta}$, we have for $1\leq j<i$,
\begin{align}
	0 <a_{i}^{(\delta T_j)} \leq a_{i}^{(\delta T_{i-1})}
	& < \frac{\sigma_i}{(\sigma_ie^{2\delta}- \lambda_i)  e^{\delta (- 2\sigma_i/\sigma_{i-1} - \eta \sigma_i^2/\sigma_{i-1})} + \sigma_i }
	\underset{\delta \to \infty}{\to} 0 \,.
\end{align}
We have $ (2  + \eta \sigma_i) \sigma_i/\sigma_{i-1} < 2$, because we assumed that $\eta < 2 \frac{\sigma_{i}- \sigma_{i+1}}{\sigma_i^2} \,,\; \forall i\quad 1\leq i \leq d$. Note that for $i= r_{xy}$ we have $a_i^{(\delta T_{i+1})} = 0 \,,\;\forall \delta >0$.

Let us now prove that  if  $j>i$, then $a_{i}^{(\delta T_j)} \underset{\delta \to \infty}{\to} \frac{\sigma_i}{\lambda_i}$.

Using~\eqref{eq:mu_asym_lowerbound} and recalling that in Thm.~\ref{thm:discrete_case}, we assume that $a_i^{(0)} = e^{-2\delta}$, we have for $j > i \geq 1$,
\begin{align}
	\frac{\lambda_i}{\sigma_i} > a_{i}^{(\delta T_j)} \geq a_{i}^{(\delta T_{i+1})}
	& > \frac{\sigma_i}{(\sigma_ie^{2\delta} - \lambda_i)  e^{\delta (- 2\sigma_i/\sigma_{i+1} + 4\eta \sigma_i^2/\sigma_{i+1})} + \sigma_i }
	\underset{\delta \to \infty}{\to} \frac{\sigma_i}{\lambda_i} \,.
\end{align}
where we have that $(e^{2\delta}-1)  e^{\delta (- 2\sigma_i/\sigma_j + 4\eta \sigma_i^2/\sigma_j)} \to 0$ because 
\begin{align*}
1 -\sigma_i/\sigma_{i+1} + 2\eta \sigma_i^2 /\sigma_{i+1}<0 
&\Leftrightarrow \eta < (\sigma_i - \sigma_{i+1})/(2\sigma_i^2) \,.
\end{align*}

Now for $i \geq r_{xy}+1$ we just need to use,~\eqref{eq:mu_sublinear} to get,
\begin{equation}
	a_i^{(t)} \leq \frac{a_i^{(0)}}{1 + \lambda \eta a_i^{(0)} t} \underset{\delta \to \infty}{\to} 0 \,.
\end{equation}

\end{document}